\documentclass{article}
\usepackage{PRIMEarxiv}

\def\mytitle{Probabilistic computation and uncertainty quantification with emerging covariance}

\usepackage[utf8]{inputenc} % allow utf-8 input
\usepackage[T1]{fontenc}    % use 8-bit T1 fonts
\usepackage{hyperref}       % hyperlinks
\usepackage{url}            % simple URL typesetting
\usepackage{booktabs}       % professional-quality tables
\usepackage{amsfonts}       % blackboard math symbols
\usepackage{nicefrac}       % compact symbols for 1/2, etc.
\usepackage{microtype}      % microtypography
\usepackage{lipsum}
\usepackage{fancyhdr}       % header
\usepackage{graphicx}       % graphics
\graphicspath{{media/}}     % organize your images and other figures under media/ folder
\usepackage{bm}

\usepackage[table,xcdraw]{xcolor}
\usepackage[normalem]{ulem}
\useunder{\uline}{\ul}{}
\usepackage{xcolor}
\usepackage{amsmath}
\usepackage{amssymb}
\usepackage{mathtools}
\usepackage{amsthm}
\usepackage{multirow}
\usepackage{caption}
\usepackage{multicol}
\newtheorem{theorem}{Theorem}

\definecolor{lightblue}{RGB}{35,123,211}

\title{\mytitle}

\def\mm{Appendix}
\def\si{Supplementary Information}
\newcommand{\ma}[1]{\textcolor{black}{#1}}%teal black
\usepackage[mathlines]{lineno}
\begin{document}
%\maketitle
\begin{flushleft}
{\Large
\mytitle
}
\newline
{
  Hengyuan Ma\textsuperscript{1},
  Yang Qi\textsuperscript{1,2,3},
  Li Zhang\textsuperscript{4},
  Jie Zhang\textsuperscript{1,2},
  Wenlian Lu\textsuperscript{1,2,5,6,7,8},
  Jianfeng Feng\textsuperscript{1,2,$\ast$} 
\\

\bigskip
\it{1} Institute of Science and Technology for Brain-inspired Intelligence, Fudan University, Shanghai 200433, China
\\
\it{2} Key Laboratory of Computational Neuroscience and Brain-Inspired Intelligence (Fudan University), Ministry of Education, China
\\
\it{3} MOE Frontiers Center for Brain Science, Fudan University, Shanghai 200433, China
\\
\it{4} School of Data Science, Fudan University, Shanghai 200433, China
\\
\it{5} School of Mathematical Sciences, Fudan University, No. 220 Handan Road, Shanghai, 200433, Shanghai, China
\\
\it{6} Shanghai Center for Mathematical Sciences, No. 220 Handan Road, Shanghai, 200433, Shanghai, China
\\
\it{7} Shanghai Key Laboratory for Contemporary Applied Mathematics, No. 220 Handan Road, Shanghai, 200433, Shanghai, China
\\
\it{8} Key Laboratory of Mathematics for Nonlinear Science, No. 220 Handan Road, Shanghai, 200433, Shanghai, China 
\newline
$\ast$ jffeng@fudan.edu.cn 
}
\end{flushleft}

\begin{abstract}
Building robust, interpretable, and secure AI system requires quantifying and representing uncertainty under a probabilistic perspective to mimic human cognitive abilities. However, probabilistic computation presents significant challenges for most conventional artificial neural network, as they are essentially implemented in a deterministic manner. In this paper, we develop an efficient probabilistic computation framework by truncating the probabilistic representation of neural activation up to its mean and covariance and construct a moment neural network that encapsulates the nonlinear coupling between the mean and covariance of the underlying stochastic network. We reveal that when only the mean but not the covariance is supervised during gradient-based learning, the unsupervised covariance spontaneously emerges from its nonlinear coupling with the mean and faithfully captures the uncertainty associated with model predictions.
Our findings highlight the inherent simplicity of probabilistic computation by seamlessly incorporating uncertainty into model prediction, paving the way for integrating it into large-scale AI systems.
\end{abstract}
keywords: Stochastic neural network, Probabilistic computation, Uncertainty quantification, Machine learning, Artificial intelligence

%\linenumbers
\section{Introduction}
Uncertainty is an inherent and omnipresent aspect of learning, arising from various sources such as the environment, observed data, learning process, and the model itself. 
While humans are capable of incorporating such uncertainty into their decision-making~\cite{kelly2015neural}, it is not easy for most of the current artificial neural networks (ANNs) to evaluate the level of uncertainty associated with their predictions~\cite{abdar2021review,gawlikowski2021survey}.
Emulating the brain's approach to information processing is one of the most viable ways towards more general artificial intelligence~\cite{goertzel2007human}.
The challenge arises from one of the fundamental distinctions between information processing in the human brain and today's hardware used for implementing ANNs: the brain employs {\em probabilistic computation} to process information~\cite{sprevak2020two}. 
In contrast to deterministic computation based on von Neumann architecture~\cite{von2012computer}, 
probabilistic computation goes beyond making predictions and enables the quantification of prediction uncertainty.
This capability is essential for more robust, trustworthy, and interpretable artificial intelligence~\cite{seuss2021bridging}.

Probabilistic computation performs a series of transformations of probability distributions to generate a desired distribution whose mean represents the prediction and whose covariance represents the prediction uncertainty associated with the relevant input. However, learning general probabilistic computations is numerically intractable due to the complexity of high-dimensional joint probability density functions~\cite{ghahramani2015probabilistic}. One of the most popular solutions to this challenge is Bayesian variational inference (BVI)~\cite{jordan1999introduction}, which employs a parameterized variational distribution and minimizes a variational loss. However, evaluating the gradient of the variational loss for this process is computationally costly. This problem is alleviated by Monte Carlo (MC) sampling methods~\cite{doucet2001sequential,paisley2012variational,kingma2013auto}.
Nevertheless, MC methods require multiple model calls or training ensembles of models in parallel, leading to increased time and space complexity compared to deterministic models.

When an ANN is used as a model in BVI, stochasticity is introduced into the network, and the mean and covariance of the network states are supervised (either explicitly or implicitly) so that the model can learn to make both predictions and to represent uncertainty~\cite{gal2016dropout,lakshminarayanan2017simple,teye2018bayesian,mobiny2021dropconnect}.
In biological neuronal systems, the mean and covariance of neuronal spiking activities exhibit nonlinear coupling~\cite{ponce2013stimulus,panzeri2022structures}, and their propagation across two populations of neurons shares the same synaptic weights~\cite{lu2010gaussian}.
This implies that when a model optimizes the output mean for a task loss, it concurrently adjusts the output covariance through this coupling.
We wonder whether this coupling can autonomously regulate the covariance, enabling it to capture prediction uncertainty without additional supervision.
This inspires us to construct a probabilistic computation framework where only the mean is supervised, without additional supervision on the covariance. This approach allows the covariance to naturally emerge through nonlinear coupling. We propose that this {\em emerging covariance} serves as a faithful representation of uncertainty.
 
Through theoretical analysis and numerical experiments we reveal two key properties of the proposed framework of probabilistic computation, one of which is that a general probabilistic computation can be approximated by another stochastic process underlied by a neural network with nonlinearly coupled covariance, and the other is that prediction uncertainty can be captured by the emerging covariance.
We develop an efficient probabilistic computation framework consisting of a model called {\em moment neural network} (MNN) and a modified gradient descent optimizer called {\em supervised mean unsupervised covariance} (SMUC).
The MNN represents a probability distribution up to its second cumulants, and conducts nonlinearly coupled mean and covariance via {\em moment activations} (MAs).
We train the MNN to learn a range of probabilistic computation tasks with SMUC, which only back-propagates the gradient through the mean but not the covariance. As learning progresses, covariance automatically emerges due to the nonlinear coupling with the mean.
By computing the prediction entropy from emerging covariance, we demonstrate exceptional capability of the MNN for uncertainty quantification including misclassified and out-of-distribution sample detection, and adversarial attack awareness. 
Moreover, our approach can achieve comparable or even superior performance for uncertainty quantification compared to existing methods.
Our approach challenges the conventional belief that uncertainty quantification requires supervision of the covariance, either explicitly or implicitly. 
Instead, we show that when the model learns making predictions, the emerging covariance faithfully captures the prediction uncertainty, suggesting that learning of probabilistic computation can be fundamentally simpler than previously believed.

\begin{figure}[t]
    \centering
    \includegraphics[width=\textwidth]{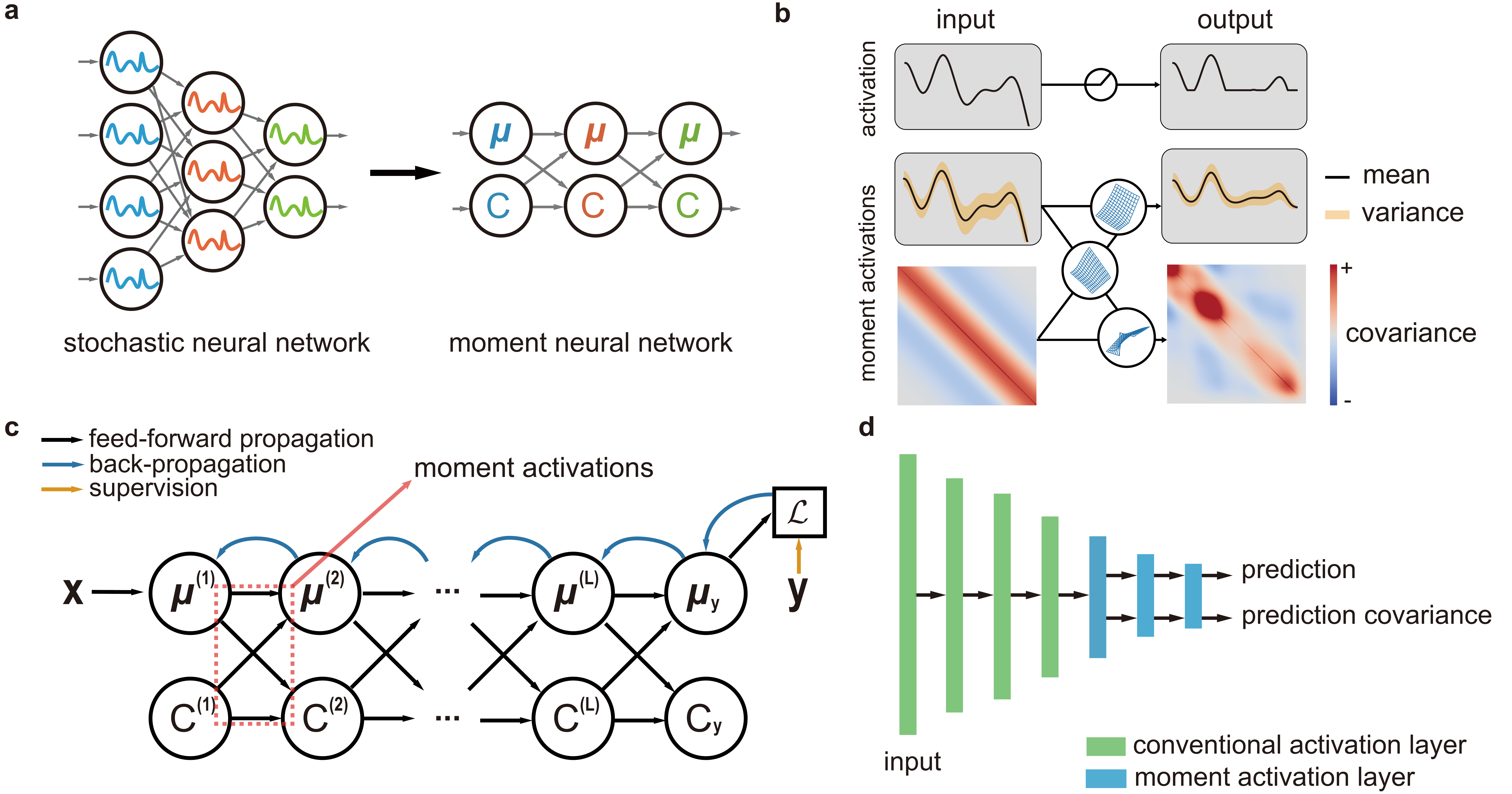}
    \caption{
    \textbf{Schematic of probabilistic computation and learning with unsupervised covariance.}
    \textbf{a}, Schematic diagram of a stochastic neural network (left) (Eq.~\eqref{eq:stoc_network}, \mm) and the corresponding moment neural network (MNN, right) (Eq.~\eqref{eq:mnn_feedforward}).
    \textbf{b},
    Comparison of conventional neural activation (top) and moment activations (MAs) (bottom). We illustrate the ReLU and ReLU MAs as examples.
    \textbf{c},
    We train MNNs using the supervised mean unsupervised covariance (SMUC) scheme, where back-propagation only involves the mean while the covariances is treated as constants. 
    \textbf{d},
    Schematics of a mixed MNN consisting of shallow layers using conventional activations and deep layers using MAs. 
    The rectangles represent the layers, and their heights indicate the respective layer dimensions.
    }
    \label{fig:frame}
\end{figure}

\section{Probabilistic computation framework for supervised learning}
Given a dataset consisting of pairs of input $\mathbf{x}\in\mathbb{R}^m$ and output (label) $\mathbf{y}\in\mathbb{R}^n$, we aim to learn a mapping from an input $\mathbf{x}$ to a probability distribution $q(\mathbf{y}|\mathbf{x})$ of the output, which incorporates both prediction and prediction uncertainty.
In general, exact calculation of such a mapping requires the knowledge of underlying generative model which is not always available. Moreover, computing such a mapping over a high-dimensional space is intractable. Here we propose that such a mapping can be approximated by a chain of simpler stages
\begin{align}\label{eq:density_chain}
    \mathbf{x}\mapsto q(\mathbf{v}^{(1)}|\mathbf{x})
              \mapsto q(\mathbf{v}^{(2)}|\mathbf{v}^{(1)})
              \mapsto \cdots
              \mapsto q(\mathbf{v}^{(L)}|\mathbf{v}^{(L-1)})
              \mapsto q(\mathbf{y}|\mathbf{v}^{(L)}),
\end{align}
where $\mathbf{v}^{(l)}$ is the $l$-th hidden state.
Equation~\eqref{eq:density_chain} can be rewritten in terms of its cumulant sequences~\cite{gardiner1985handbook}
\begin{align}\label{eq:moment_chain}
    \mathbf{x}\mapsto \{m_z^{(1)}\} 
    \mapsto \{m_z^{(2)}\}
    \mapsto \cdots
    \mapsto \{m_z^{(L)}\}
    \mapsto \{m_{z,y}\},
\end{align}
where $m_z^{(l)}$ represents the $z$-th order cumulants of the the distribution %$q(\mathbf{v}^{(l)}|\mathbf{v}^{(l-1)})$, 
$q(\mathbf{v}^{(l)}|\mathbf{x})=\int q(\mathbf{v}^{(l)}|\mathbf{v}^{(l-1)}) p(\mathbf{v}^{(l-1)}|\mathbf{x})d \mathbf{v}^{(l-1)}$, with $z=1,2,\ldots,+\infty$.
When higher-order cumulants beyond $z=Z$ vanish, the cumulant sequences can be truncated by considering only the first $Z$ cumulants for each distribution.
When $Z=1$, the truncated process is written as
\begin{align}\label{eq:deter_chain}
     \mathbf{x}\mapsto \bm{\mu}^{(1)}
    \mapsto \bm{\mu}^{(2)}
    \mapsto \bm{\mu}^{(3)}
    \mapsto \cdots
    \mapsto \bm{\mu}^{(L)}
    \mapsto \bm{\mu}_{y},
\end{align}
where $\bm{\mu}^{(l)}$ is the mean of the distribution $q(\mathbf{v}^{(l)}|\mathbf{x})$, and $\bm{\mu}_{y}$ is the mean of model prediction.
We call Eq.~\eqref{eq:deter_chain} as {\em deterministic computation}, since the output $\bm{\mu}_y$ is deterministic and the prediction uncertainty is always zero, and the truncated process at $Z\geq 2$ as {\em probabilistic computation}, where second- or higher- order cumulants account for the uncertainty of the process.
When $Z=2$, the truncated process can be interpreted as a Gaussian approximation, where the transition of the mean $\bm{\mu}^{(l)}$ and covariance $C^{(l)}$ across each layer can be expressed as
\begin{align}\label{eq:trunc}
    \mathbf{x}\mapsto \{\bm{\mu}^{(1)},C^{(1)}\}
    \mapsto \{\bm{\mu}^{(2)},C^{(2)}\}
    \mapsto \cdots
    \mapsto \{\bm{\mu}^{(L)},C^{(L)}\}
    \mapsto \{\bm{\mu}_{y},C_y\},
\end{align}
with $C_y$ representing the prediction covariance. 
When $3\leq Z<\infty$, the truncated process cannot be regarded as a stochastic process, rendering the entropy undefined. This is due to that the cumulant sequences of a probability distribution either has all but the first two cumulants vanish or has an infinite number of non-vanishing cumulants~\cite{gardiner1985handbook}. 
Therefore, we focus on the $Z=2$ case as this is the minimal model that is capable of representing uncertainty. 
The remaining issue is to develop a learnable model for effectively conducting the truncated process and to devise a suitable training approach for this model.
For deterministic computation (Eq~\eqref{eq:deter_chain}), this can be achieved by conventional ANNs via compositions of linear and nonlinear transformations with trainable weights and biases. Due to its universal approximation property~\cite{pinkus1999approximation}, an ANN can learn to approximate arbitrary continuous functional mapping. In the following, we show that a generalized neural network incorporating second-order cumulants can be used to learn probabilistic computation.
\section{Training stochastic networks with emerging covariance}
\subsection{\ma{Moment activations (MAs) and moment neural networks (MNNs)}}

We propose an extension of conventional ANNs by incorporating stochasticity in order to approximate mappings from vector spaces to probability spaces beyond simple functions between vector spaces.

Consider a multi-layer stochastic neural network as follows
\begin{align}\label{eq:stoc_network}
\begin{aligned}
   \frac{d \mathbf{x}^{(1)}}{dt} &= -\mathbf{x}^{(1)} + \mathbf{x} +  \sqrt{2}\sigma_1\bm{\xi}^{(1)},
    \quad\mathbf{v}^{(1)} =\mathbf{x}^{(1)}\\
    \frac{d \mathbf{x}^{(l)}}{dt} &= -\mathbf{x}^{(l)} + W^{(l-1)}\mathbf{v}^{(l-1)} + \mathbf{b}^{(l-1)}+ \sqrt{2}\sigma_l\bm{\xi}^{(l)},
    \quad\mathbf{v}^{(l)} = \mathbf{h}(\mathbf{x}^{(l)}),\quad l=2,\ldots,L\\
    \frac{d \mathbf{y}}{dt} &= -\mathbf{y} + W^{(L)}\mathbf{v}^{(L)},
    \end{aligned}
\end{align}
where \ma{$\mathbf{x},\mathbf{y}$ are the input and output of the system, $\mathbf{x}^{(l)}$ is the neural state of layer $l$ corresponding to one stage in Eq.~\eqref{eq:density_chain}}, $\mathbf{v}^{(l)}$ is the signal transmitted to neurons at the next layer, $W^{(l)}, \mathbf{b}^{(l)}$ are learnable weights and biases, $\bm{\xi}^{(l)}$ is the unit Gaussian white noise, $\sigma_{l}^2$ is constant diffusion coefficients, and $\mathbf{h}(\cdot)$ is an element-wise nonlinearity which transfers each coordinate $x_i$ through a real-valued function $h_i(x_i)$.
The goal is to train this stochastic network to approximate $\mathbf{x} \mapsto q(\mathbf{y}|\mathbf{x})$.
Directly training requires MC methods for estimating the gradient of the network which is computationally costly.
{Instead, we approximate the system by constructing its corresponding truncated process up to the first two cumulants (Eq.~\eqref{eq:trunc}. To do so, we assume that the system has reached its stationary distribution, then estimate the stationary mean and covariance $\{\bm{\mu}^{(l)},C^{(l)}\}$ of the signal vector $\mathbf{v}^{(l)}$ for $l=1,\ldots,L$ as well as the output mean and covariance $\{\bm{\mu}_y,C_y\}$.
For the first layer, the mean and covariance can be directly calculated from Eq.~\eqref{eq:stoc_network}}
\begin{align}\label{eq:first_layer}
    \bm{\mu}^{(1)} = \mathbf{x},\quad C^{(1)}=\sigma_1^2I.
\end{align}
We also have the relationship between the cumulants of the last two layers:
\begin{align}\label{eq:last_two_layer}
\bm{\mu}_{y}=W^{(L)}\bm{\mu}^{(L)},\quad C_{y}=W^{(L)}C^{(L)}(W^{(L)})^{\mathrm{T}}.
\end{align}
For other layers, it is unfeasible to calculate the cumulants of the signal $\mathbf{v}^{(l)}$ due to the nonlinearity $\mathbf{h}(\cdot)$. Nevertheless, we can estimate the truncated transition $\{\bm{\mu}^{(l-1)},C^{(l-1)}\} \mapsto \{\bm{\mu}^{(l)},C^{(l)}\}$ between layers using a mean field approximation (see Sec.1-2, \si~for details)
\begin{align}\label{eq:layer_ma}
\bm{\mu}^{(l)}=\mathbf{m}_v(\bar{\bm{\mu}}^{(l)},\bar{C}^{(l)}),\quad
    C^{(l)}=C_v(\bar{\bm{\mu}}^{(l)},\bar{C}^{(l)}),\quad l=1,\ldots,L,
\end{align}
where $\bar{\bm{\mu}}^{(l)}=W^{(l-1)}\bm{\mu}^{(l-1)}+\mathbf{b}^{(l-1)}$ and $\bar{C}^{(l)}=W^{(l-1)}C^{(l-1)}(W^{(l-1)})^{\mathrm{T}}+\sigma_l^2I$ are the linear summations of mean and covariance according to the weights and biases. The two nonlinear mappings $\mathbf{m}_v(\cdot,\cdot)\in\mathbb{R}^n\times\mathbb{R}^{n\times n}\rightarrow \mathbb{R}^n$ and $C_v(\cdot,\cdot)\in\mathbb{R}^n\times\mathbb{R}^{n\times n}\rightarrow \mathbb{R}^{n\times n}$ are {\em moment activations (MAs)}
\begin{align}
    &\mathbf{m}_{v}(\bm{\bar{\mu}},\bar{C})_i = m_{v}(\bar{\mu}_i,\bar{C}_i),\quad 1\leq i \leq n\label{eq:mean_map}\\
    &C_{v}(\bm{\bar{\mu}},\bar{C})_{ij}= \left\{\begin{array}{@{}ll@{}}
     C_{v}(\bar{\mu}_i,\bar{C}_i),&\quad 1\leq i=j\leq n\\
     \chi(\bar{\mu}_i,\bar{C}_i)\chi(\bar{\mu}_j,\bar{C}_j)\tfrac{\bar{C}_{ij}}{\sqrt{\bar{C}_i\bar{C}_j}},&\quad 1\leq i\neq j \leq n,
    \end{array}\right.\label{eq:cov_map}
\end{align}
where the dimension of the $\bm{\bar{\mu}}^{(l)}$ is $n$, and three nonlinear functions $m_v(\cdot,\cdot),C_v(\cdot,\cdot)$ and $\chi(\cdot,\cdot)$ can be derived for concrete 
activation $h(\cdot)$.
MAs encapsulate the nonlinear transformation from the mean and covariance of the input to the output at each layer, as illustrated in Fig.~\ref{fig:frame}\textbf{b}.
Combining Eq.~\eqref{eq:first_layer}-\eqref{eq:layer_ma}, the stochastic neural network (Eq.~\eqref{eq:stoc_network}) now is reduced to the following nonlinear deterministic network
\begin{align}
\begin{aligned}\label{eq:mnn_feedforward}
\begin{aligned}
    &\bm{\mu}^{(1)}=\mathbf{x},\\
    &\begin{array}{@{}ll@{}}
    \bar{\bm{\mu}}^{(l)}=W^{(l-1)}\bm{\mu}^{(l-1)}+\mathbf{b}^{(l-1)},\\
\bm{\mu}^{(l)}=\mathbf{m}_v(\bar{\bm{\mu}}^{(l)},\textcolor{lightblue}{\bar{C}^{(l)}}),
    \end{array}\\
    &\bm{\mu}_{y}=W^{(L)}\bm{\mu}^{(L)},
\end{aligned}
\begin{aligned}
&\textcolor{lightblue}{C^{(1)}=\sigma_1^2I},\\
    &\left.\begin{array}{@{}ll@{}}
    \textcolor{lightblue}{\bar{C}^{(l)}=W^{(l-1)}C^{(l-1)}(W^{(l-1)})^{\mathrm{T}}+\sigma_l^2I},\\
    \textcolor{lightblue}{C^{(l)}=C_v(\bar{\bm{\mu}}^{(l)},\bar{C}^{(l)})},
    \end{array}\right\}  l=2,\ldots,L\\
&\textcolor{lightblue}{C_{y}=W^{(L)}C^{(L)}(W^{(L)})^{\mathrm{T}}}.
\end{aligned}
\end{aligned}
\end{align}
We call the network Eq.~\eqref{eq:mnn_feedforward} as {\em moment neural networks (MNN)}.
\ma{The MNN effectively calculates the mean and covariance of the signals $v^{(l)}$ in each layer of the stochastic neural network (Eq.~\eqref{eq:stoc_network}). In this way, the MNN can estimate the output mean and covariance of the stochastic neural network, without conducting its stochastic dynamics,} as illustrated in Fig.~\ref{fig:frame}\textbf{a}.
When the covariances (blue part in Eq.~\eqref{eq:mnn_feedforward}) are fixed at zero, the MNN reduces to the conventional ANNs.

\subsection{{Supervised mean unsupervised covariance (SMUC)}}

Suppose a specific configuration ($W^{(l)},\mathbf{b}^{(l)},\sigma_l$) exists that enables the stochastic network Eq.~\eqref{eq:stoc_network} well fits the mapping $\mathbf{x} \rightarrow q(\mathbf{y}|\mathbf{x})$ of the dataset. We refer to this network as the ground-truth model, and denote the corresponding MNN as $M_{GT}$. We aim to train an MNN denoted as $\tilde{M}$ with the same architecture as Eq.~\eqref{eq:stoc_network} to fit $M_{GT}$, enabling both prediction and uncertainty quantification. Since $\tilde{M}$ contains means and covariances of each layer, the conventional perspective for training $\tilde{M}$ necessitates the supervision of both of them.
This involves incorporating both mean and covariance in the loss function and backpropagating the error gradient through the mean and covariance of each layer during training. 
We call this training approach as {\em supervised mean supervised covariance} (SMSC).
One drawback of SMSC is the computational cost for backpropagating through covariances compared to ANNs involving only the mean, \ma{as it requires the calculation and storage the gradient of a quadratic number of elements for each layer.}

Nevertheless, in many neural systems, both biological and artificial alike, the mean and covariance of neural activity are not independent but nonlinearly coupled~\cite{ponce2013stimulus,panzeri2022structures}. 
We wonder whether this nonlinear coupling can be exploited to autonomously regulate the covariance. Based on this intuition, we propose a modified gradient descent algorithm called {\em supervised mean unsupervised covariance (SMUC)} for training the MNN, which only backpropagates the error gradient through the mean while treating the gradients with respect to covariances (blue part in Eq.~\eqref{eq:mnn_feedforward}) as zero.
Concretely, given input-output pairs $(\mathbf{x},\mathbf{y})$ from the dataset $\mathcal{D}$, the task loss function $\mathcal{L}$, and learning rate $\gamma_t$, the SMUC updates the parameters $\theta=\{W^{(l)},\mathbf{b}^{(l)}\}$ of the MNN as follows
\begin{align}\label{eq: update_theta}
    \theta_{t+1} = \theta_{t}- \gamma_{t} \sum_{(\mathbf{x},\mathbf{y})\in\mathcal{D}} \frac{\breve{\partial}  \mathcal{L}(\mathbf{y},\bm{\mu}_{y}(\mathbf{x},\theta_{t}))}{\breve{\partial}   \theta_{t}},
\end{align}
where $\breve{\partial}$ is a modified gradient operator. When calculating the gradient using $\breve{\partial}$, we backpropagate the error gradient through the chain rule only on the mean while treating the covariances as constants:
\begin{align}
\begin{aligned}
    &\frac{\breve{\partial}  \mathcal{L}}{\breve{\partial}   \theta}= \frac{\partial  \mathcal{L}}{\partial   \bm{\mu}^{(L)}}\frac{\breve{\partial}  \bm{\mu}^{(L)}}{\breve{\partial}  \theta},\\
&
\frac{\breve{\partial}  \bm{\mu}^{(l)}}{\breve{\partial}  \theta} =  \frac{\partial \bm{\mu}^{(l)}}{\partial  \bm{\mu}^{(l-1)}}\frac{\breve{\partial}  \bm{\mu}^{(l-1)}}{\breve{\partial}   \theta} +
 \frac{\partial C^{(l)}}{\partial  C^{(l-1)}}\frac{\breve{\partial} C^{(l-1)}}{\breve{\partial}  \theta},\quad\forall l=L,\ldots,2,\quad \frac{\breve{\partial}  \bm{\mu}^{(1)}}{\breve{\partial}  \theta} = \frac{{\partial}  \bm{\mu}^{(1)}}{{\partial}  \theta},\\
&\frac{\breve{\partial}  C^{(l)}}{\breve{\partial}  \theta} = 0,\quad\forall l=L,\ldots,1,
\end{aligned}
\end{align}
A schematic diagram for this learning rule is illustrated in Fig.~\ref{fig:frame}\textbf{c}. 
The direct consequence of this approach is that only the mean is supervised for minimizing the loss, while the covariance spontaneously emerges through the nonlinear coupling with the mean during forward computation.

\subsection{{Theoretical analysis of the SMUC}}

We analyze the theoretical properties of the SMUC to demonstrate that the nonlinearly coupling between the mean and covariance helps to capture the uncertainty of the model. We find that SMUC is equivalent to stochastic Riemannian gradient descent (SRGD)~\cite{bonnabel2013stochastic}, a variant of the stochastic gradient descent (SGD) algorithm.
At each iteration of the SRGD, an SGD update is conducted on the parameter, which is then projected onto a Riemannian manifold. 
In our SMUC, it turns out that at each iteration, the parameter $\theta$ is projected onto the Riemannian manifold where the variance at each layer are identical to that of the ground-truth MNN $M_{GT}$. (See \si~for more details). Based on this, we further prove the convergence of the SMUC under mild conditions as shown in Thm.~\ref{thm:rgd}.
\begin{theorem}\label{thm:rgd}
    If the learning rate schedule $\{\gamma_t\}$ satisfies a modified usual condition (Eq.~\eqref{eq:usu_cond}, \si), the parameter $\theta$ converges to a point at which the gradient diminishes almost surely.
\end{theorem}
Next, we investigate what the MNN can learn from the dataset after convergence. We prove that the MNN learns the mean and variance at each layer of $M_{GT}$ after convergence, as shown in Thm.~\ref{thm:mean_var}.
\begin{theorem}\label{thm:mean_var}
    After convergence, $\tilde{M}$ learns the mean and variance of $M_{GT}$ at each layer.
\end{theorem}
For the covariance, the situation is more complicated. Although there can be error on the covariance at each layer of $\tilde{M}$, we prove that for sufficiently deep networks, the output covariance of $\tilde{M}$ converges to that of $M_{GT}$, even when the input covariance is randomly chosen, as shown in Thm.~\ref{thm:cov}.
\begin{theorem}\label{thm:cov}
After convergence, denote $\lambda_{\min}(C^{(l)})$ as the smallest eigenvalue of $C^{(l)}$, and define the {\em covariance deviation rate} at the $l$-th layer 
\begin{align}\label{eq:dev_rate}
    r^{(l)} = \frac{n^{(l)}(\max_{i}\chi_i^{(l)})^2}{\sqrt{\lambda_{\min}(C^{(l)})^2+2\lambda_{\min}(C^{(l)})\sigma_l^2}}
\end{align}
with 
\begin{align}
    \chi_i^{(l)} = \chi(W_{i\bullet}^{(l)}\mathbf{v}^{(l)},W_{i\bullet}^{(l)}C^{(l)}(W_{i\bullet}^{(l)})^{\mathrm{T}}+\sigma_l^2).
\end{align}
Then, as the total number of layers approaches infinity, the output covariance of the $\tilde{M}$ converges to the output covariance of $M_{GT}$, given that the series $-\sum_{l} \log r^{(l)}$ diverges to positive infinity. 
\end{theorem}

As a result, the MNN's ability of uncertainty quantification is maintained through the output covariance $C_{y}$ without tracking of covariances in the shallow layers.
This allows us to construct a hybrid model called {\em mixed MNN} (Fig.~\ref{fig:frame}\textbf{d}), where we set the covariance in the shallow layers of the network to zero, reducing them to a conventional ANNs, whereas only in the last few layers the covariance are computed.

Since the dimensions of the deep layers are lower, the mixed MNN significantly reduces the computational cost on calculating and storing the covariance matrices.
See \si~for the proof of Thm.~\ref{thm:rgd}-\ref{thm:cov}.
Furthermore, the complexity for training MNNs can be further reduced by sharing the covariances within a mini-batch (\mm).

\subsection{{Instantialization of the MNNs}}

For cases where the activations of the stochastic network are Heaviside function
\begin{align}
    \Theta(x) = \begin{cases}
     1,\quad x\geq 0\\
    0,\quad x<0
    \end{cases}
\end{align}
and ReLU function
 \begin{align}
    \mathrm{ReLU}(x) = \begin{cases}
     x,\quad x\geq 0\\
    0,\quad x<0
    \end{cases},
\end{align}
we derive expressions of $m_v(\cdot,\cdot)$, $C_v(\cdot,\cdot)$, and $\chi(\cdot,\cdot)$ in the corresponding MAs (Eq.~\eqref{eq:mean_map}-\eqref{eq:cov_map})
as follows 

\noindent\textbf{Heaviside function. }
\begin{align}
     &m_{v}(\bar{\mu},\bar{C}) =  \frac{1}{2} + \frac{1}{\sqrt{2\pi}}\int_{0}^{\frac{\bar{\mu}}{\sqrt{\bar{C}}}}e^{-\frac{x^2}{2}}dx,\label{eq:heav_mapping}\\
   &C_{v}(\bar{\mu},\bar{C}) = \frac{1}{2} + \frac{1}{\sqrt{2\pi}}\int_{0}^{\frac{\bar{\mu}}{\sqrt{\bar{C}}}}e^{-\frac{x^2}{2}}dx-(\frac{1}{2} + \frac{1}{\sqrt{2\pi}}\int_{0}^{\frac{\bar{\mu}}{\sqrt{\bar{C}}}}e^{-\frac{x^2}{2}}dx)^2,\label{eq:variance_mapping}\\
    &\chi(\bar{\mu},\bar{C}) = \frac{1}{\sqrt{2\pi}}\exp(-\frac{\bar{\mu}^2}{2\bar{C}}).\label{eq:heaviside_chi}
\end{align}

\noindent\textbf{ReLU function. }
\begin{align}
    &m_{v}(\bar{\mu},\bar{C}) 
     =\frac{\sqrt{\bar{C}}}{\sqrt{2\pi}}\exp(-\frac{\bar{\mu}^2}{2\bar{C}})+\bar{\mu}(\frac{1}{2}+\frac{1}{\sqrt{2\pi}}\int_0^{\frac{\bar{\mu}}{\sqrt{\bar{C}}}}e^{-\frac{x^2}{2}}dx),\\
    &C_{v}(\bar{\mu},\bar{C}) =(\bar{C}+\bar{\mu}^2)(\frac{1}{2}+\frac{1}{\sqrt{2\pi}}\int_0^{\frac{\bar{\mu}}{\sqrt{\bar{C}}}}e^{-\frac{x^2}{2}}dx)+\frac{\bar{\mu}\sqrt{\bar{C}}}{\sqrt{2\pi}}\exp(-\frac{\bar{\mu}^2}{2\bar{C}})-m_{v}(\bar{\mu},\bar{C})^2,\\
    &\sqrt{\bar{C}}(\frac{1}{2}+\frac{1}{\sqrt{2\pi}}\int_{0}^{\frac{\bar{\mu}}{\sqrt{\bar{C}}}}e^{-\frac{x^2}{2}}dx).\label{eq:relu_chi}
\end{align}
We also consider MAs derived from a type of biological spiking neuron model, called the leaky integrate-and-fire (LIF) neuron model~\cite{lu2010gaussian}, 
and corresponding MNN dubbed {\em LIF MNN}, see Eq.~\eqref{eq:mu_lif}-\eqref{eq:cov_lif} in \mm~for details.

\begin{figure}[h]
    \centering
    \includegraphics[width=\textwidth]{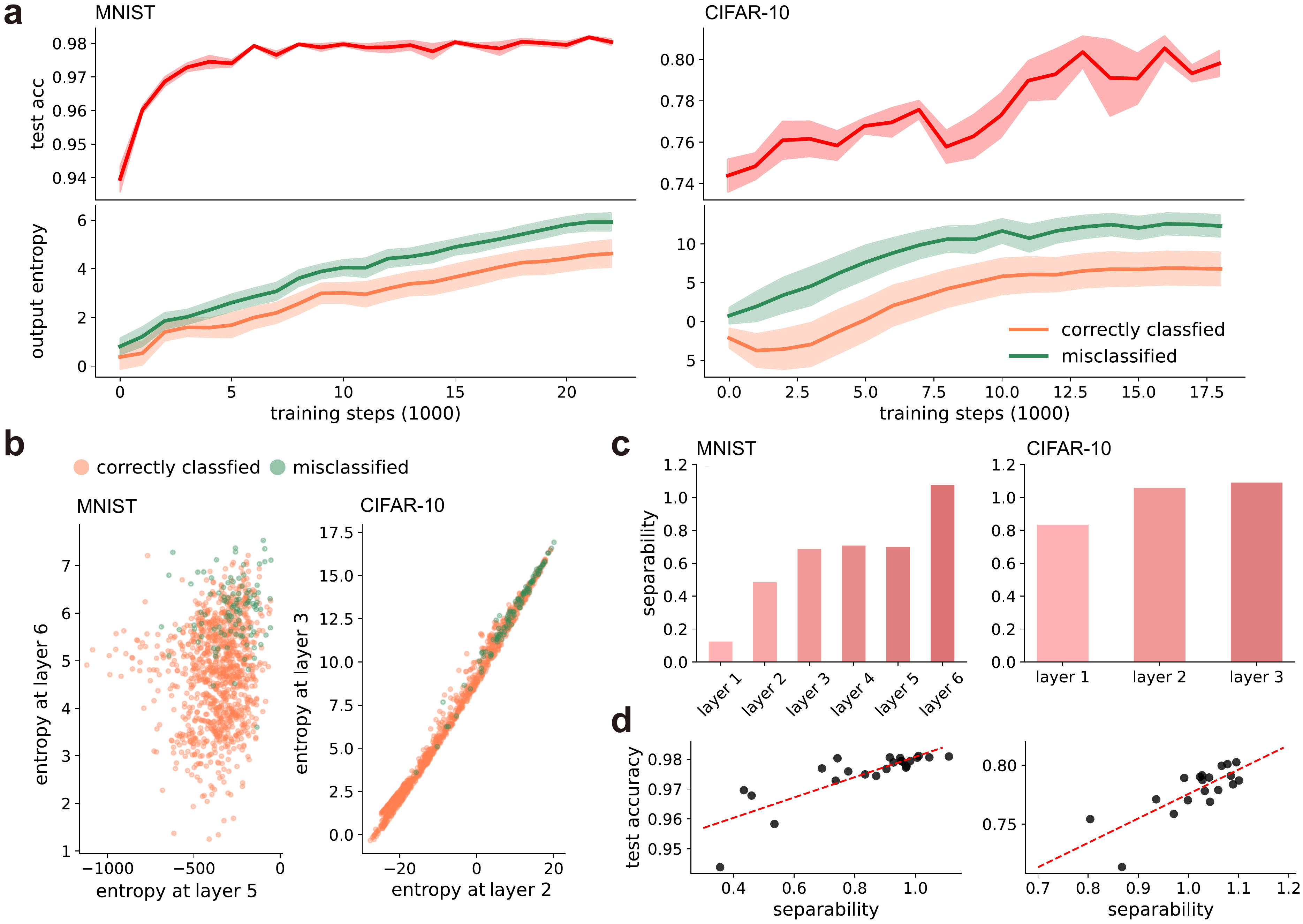}
    \caption{{\bf Emerging covariance faithfully captures prediction uncertainty}. A ReLU MNN and a mixed ReLU MNN are trained on MNIST and CIFAR-10 datasets for image classification respectively. 
    \textbf{a},
    Upper panel: the test accuracy during training. Shades indicate half standard deviation across trails; lower panel: the entropy for correctly classified (orange) and misclassified (green) inputs on the test set as learning progresses. Shades indicate half standard deviation across input samples.
    \textbf{b},
    The entropy of the network state in the last two layers of the network for correctly classified and misclassified inputs.  
    \textbf{c},
    Separability of entropy between correctly classified and misclassified inputs increases with the layer index of the MNN. 
    \textbf{d},
    Separability and test accuracy during training are positively correlated. Dashed lines represent linear regression, with coefficients of determination found to be 0.6803 for MNIST and 0.6068 for CIFAR-10.
    }
    \label{fig:classifier_all}
\end{figure}

\subsection{{Numerical verification}}

\begin{figure}[h]
    \centering
    \includegraphics[width=0.95\textwidth]{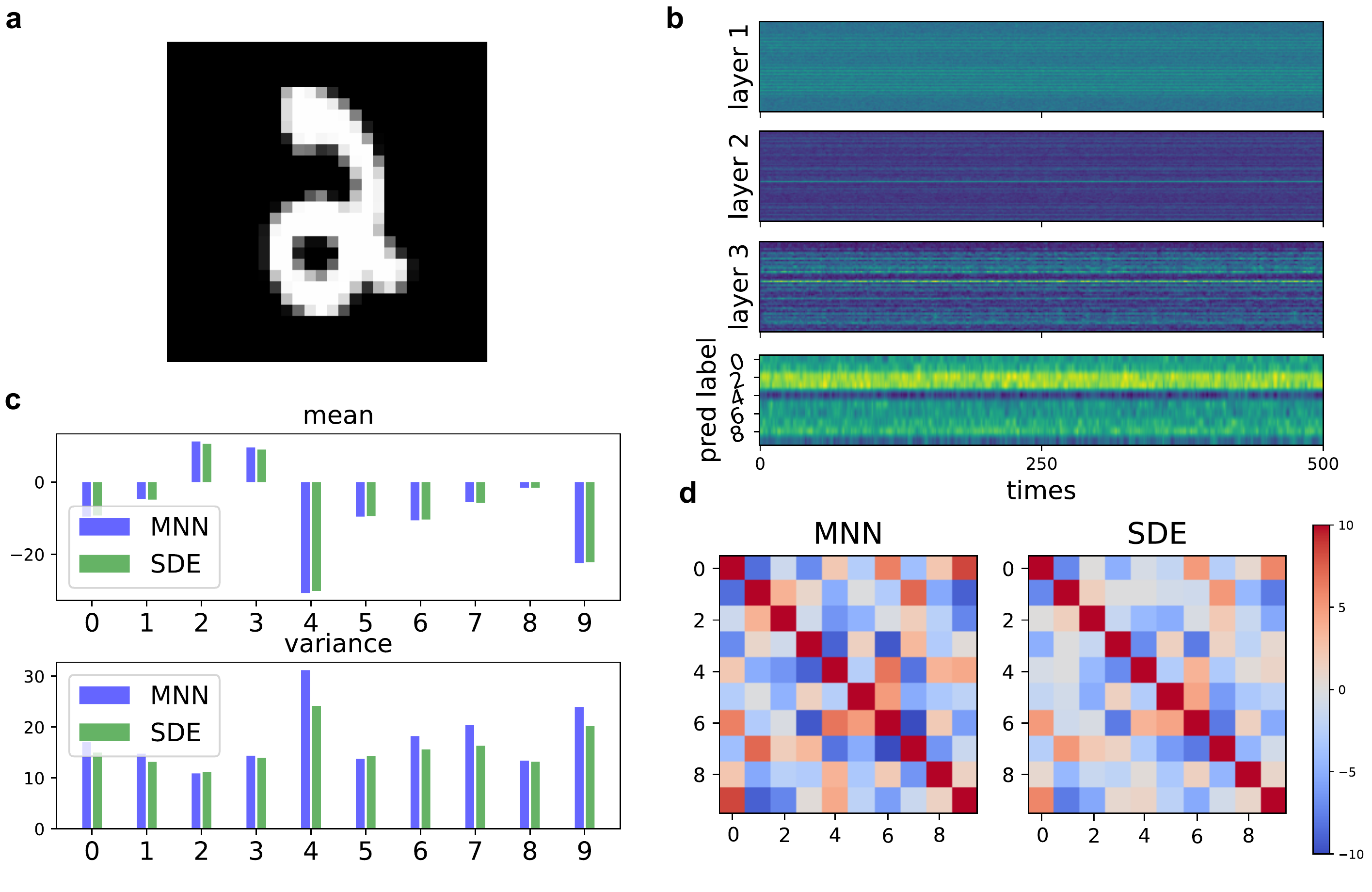}
\caption{\textbf{The ReLU MNN (Eq.~\eqref{eq:mnn_feedforward}) well approximates the corresponding stochastic network (Eq.~\eqref{eq:stoc_network}) up to its first two cumlants.}
{\bf a,} The input data from MNIST.
{\bf b,} We train an MNN on MNIST and use the parameters of the trained MNN to construct the corresponding stochastic network, and show its dynamics. 
{\bf c,} The mean and variance produced by MNN and estimated using Monte Carlo methods from the stochastic network. 
{\bf d,} The covariance matrix produced by MNN and estimated using Monte Carlo methods from the stochastic network.
}
    \label{fig:verification1}
\end{figure}

We numerically verify that the MNN (Eq.~\eqref{eq:mnn_feedforward}) well approximates the output mean and covariance of the corresponding stochastic network (Eq.~\eqref{eq:stoc_network}) in Fig.~\ref{fig:verification1}, \mm.
We train a fully connected ReLU MNN to perform image classification on the MNIST hand-written digit dataset~\cite{krizhevsky2009learning}, then use the trained parameters to construct the corresponding stochastic neural network (Eq.~\eqref{eq:stoc_network}). When provided with an input (Fig.~\ref{fig:verification1}\textbf{a}), the stochastic network generates corresponding dynamic patterns. (Fig.~\ref{fig:verification1}\textbf{b}). We simulate the stochastic neural network using the Euler-Maruyama method~\cite{gardiner1985handbook}, calculate sample estimates of its output mean, variance and covariance, and compared them with the output of the MNN (Fig.~\ref{fig:verification1}\textbf{c},\textbf{d}). 
It is observed that MNN approximates the stochastic network well in terms of the output mean and covariance. We conduct the same experiment for Heaviside MNN and obtain consistent results, as shown in Fig.~\ref{fig:verification2}, \mm.

\section{Emerging covariance faithfully captures uncertainty}

Having provided the essential elements of our theoretical framework (the MNN and the SMUC learning algorithm), we now turn to training MNN with SMUC to quantify uncertainty on various tasks. For classification task, we demonstrate the MNN's ability for quantifying uncertainty by comparing its performances on correctly classified, misclassified, out-of-distribution and adversarially attacked samples. For regression task, we evaluate the performance of MNN through the log-likelihood measurement.

\subsection{{In-distribution prediction uncertainty}}
We train a fully connected, multi-layered ReLU MNN on MNIST~\cite{lecun1998gradient} and a mixed ReLU MNN on CIFAR-10~\cite{krizhevsky2009learning} datasets for classification.
The prediction is determined by the output mean $\mathbf{m}_y$, which produces accuracy comparable to ANNs' as shown in Fig.~\ref{fig:classifier_all}\textbf{a}. However, unlike ANNs, the MNN also expresses prediction uncertainty through the output covariance $C_y$. To quantify the prediction uncertainty, we calculate the entropy of the prediction distribution $\mathcal{N}(\mathbf{m}_y,C_y)$, which depends solely on the output covariance $C_y$ (\mm).
As the training progresses, the model exhibits higher entropy on misclassified inputs compared to correctly classified ones (Fig.~\ref{fig:classifier_all}\textbf{a}), indicating that the model can effectively express uncertainty in its output covariance.
We assess the statistical disparity in the entropy across layers between correctly classified and misclassified inputs by calculating their {separability} (\mm), which quantifies the degree of separation between network responses to correctly classified and misclassified inputs. As the layer index increases, the separability also increases (Fig.\ref{fig:classifier_all}\textbf{c}), indicating an improved quality of uncertainty representation with deeper layers. 
While it is widely acknowledged that increasing the depth of a neural network improves its fitting ability~\cite{telgarsky2016benefits}, our finding provides an additional perspective: deeper networks enable more elaborate representation of uncertainty.
We also find a considerably positive correlation between the separability and accuracy on the test set throughout the training procedure (Fig.~\ref{fig:classifier_all}\textbf{d}), suggesting that the performance of prediction and uncertainty quantification are learned concurrently, even without explicit supervision on covariance.
\ma{Consistent results are observed in the Heaviside MNN (Fig.~\ref{fig:classifier_all_heav}, \mm).}

\begin{figure}[h]
    \centering
    \includegraphics[width=0.7\textwidth]{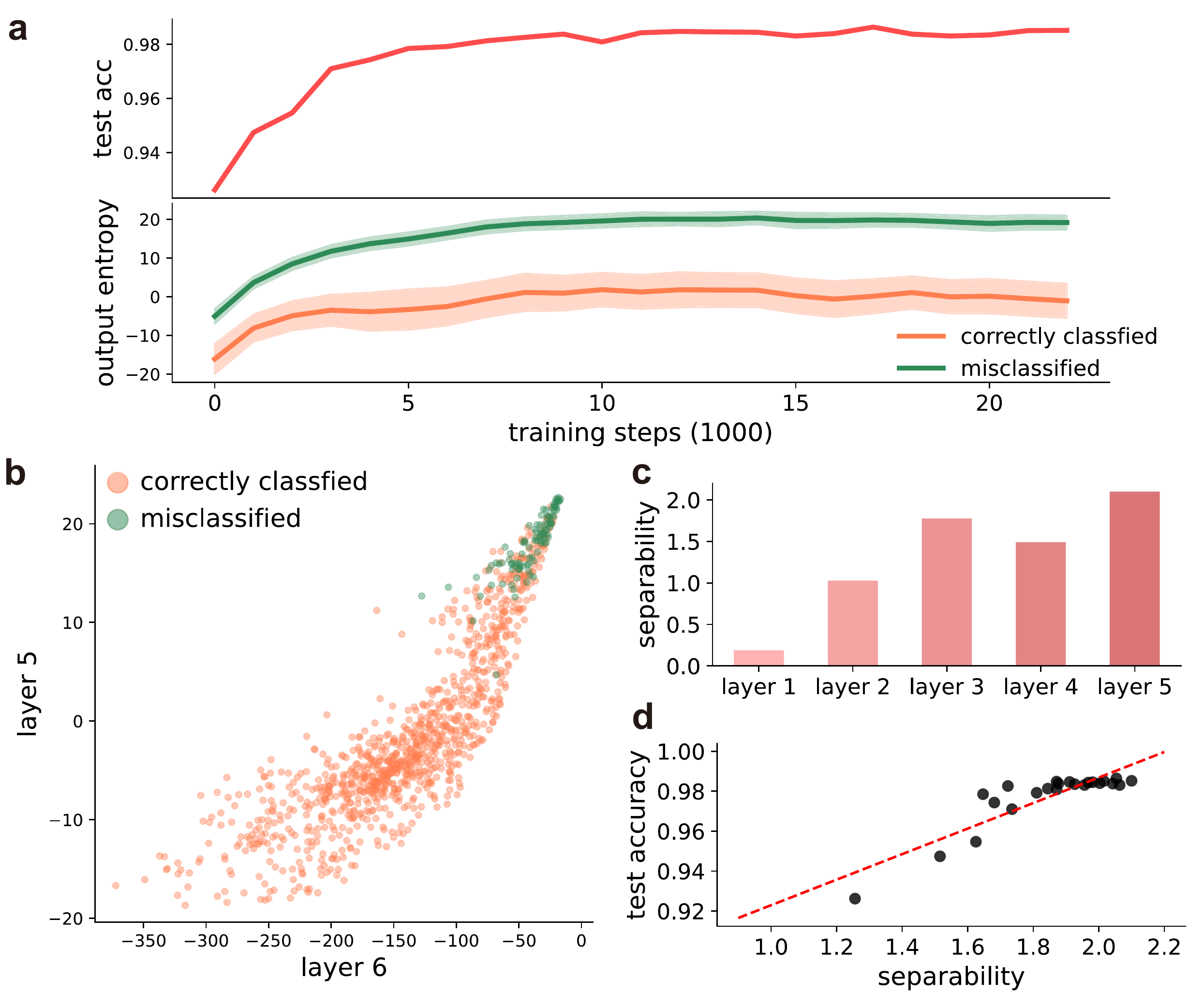}
    \caption{{\bf Emerging covariance of leaky integrate-and-fire (LIF) MNN faithfully captures the prediction uncertainty. }LIF MNN trained on MNIST for image classification.
    \textbf{a,}
    The training of MNNs on MNIST, with the accuracy (red line) and entropy (orange and green for correctly classified inputs and misclassified inputs respectively) on the test set.
    %
    %The shadows indicate the level of half standard deviations. As the training progresses, the entropy of correctly classified inputs and misclassified inputs diverges.
    \textbf{b,}
    The entropy of correctly classified and misclassified inputs in the last two layers of MNIST. The misclassified inputs result in relatively higher entropy in both layers.
    \textbf{c,}
    The entropy separability of correctly classified and misclassified inputs on MNIST increases as the layer index increases.
    \textbf{d,}
    The relationship between the separability and test accuracy during training on MNIST was analyzed using linear regression. 
    %The red dashed line represents the regression line, 
    The with coefficient of determination of 0.7760.
    }
    \label{fig:classifier_lif_all}
\end{figure}

\ma{We further investigate if the emerging covariance of the LIF MNN can also exhibit high uncertainty on the misclassified samples compared to the correctly classified samples. As shown in Fig.~\ref{fig:classifier_lif_all}, we train a LIF MNN using SMUC, and observe the consistent results. This implies that the LIF network can also utilize the nonlinear coupling of the mean and covariance for efficient uncertainty quantification. Our finding sheds light on the mechanism through which biological neural systems capture their prediction uncertainty based on the stochastic neural fluctuation. 
}

\subsection{{Out-of-distribution  prediction uncertainty}}

\begin{figure}[h]
    \centering
    \includegraphics[width=\textwidth]{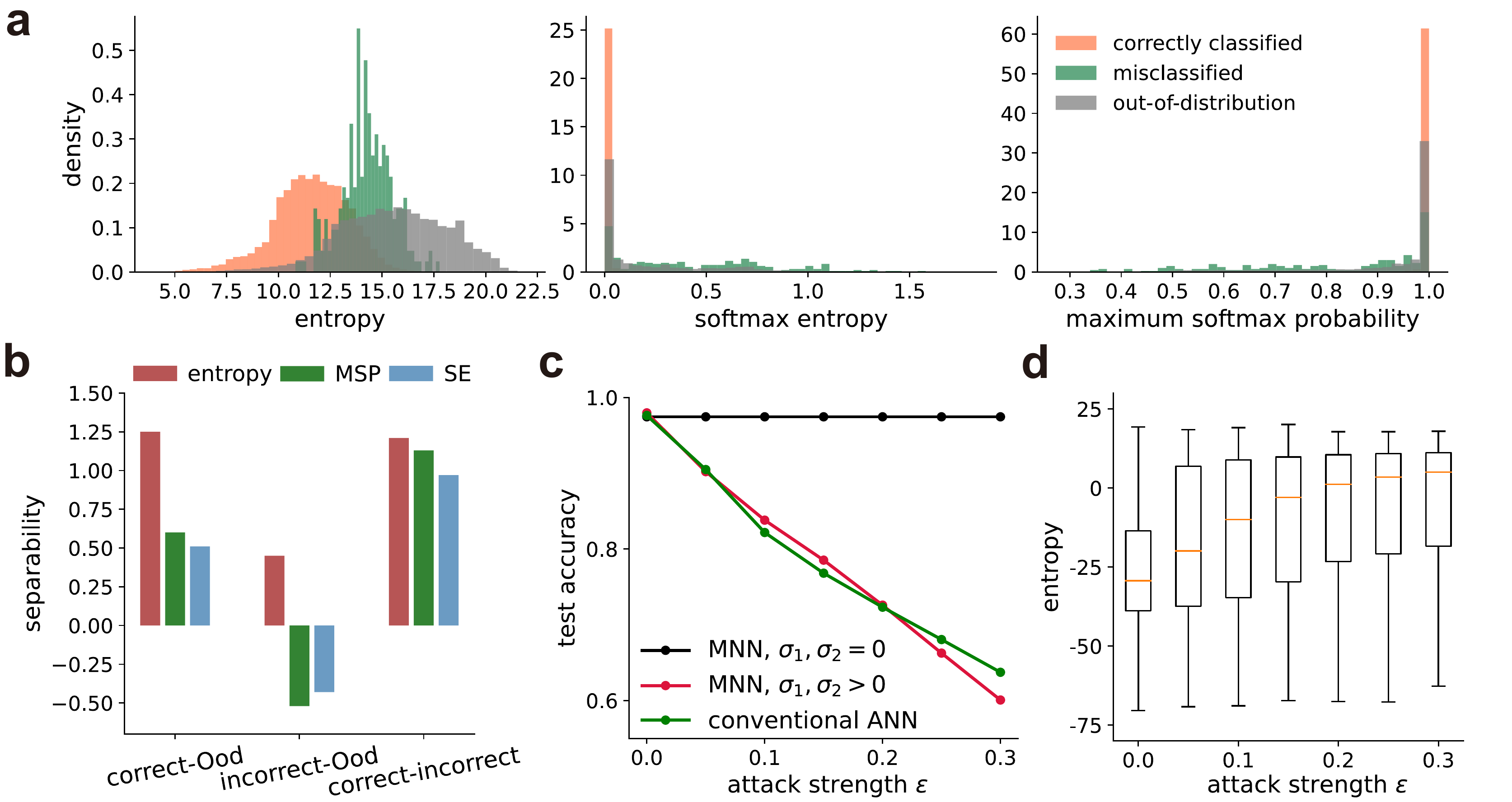}
    \caption{\textbf{
    Out-of-distribution detection and adversarial attacks awareness in the MNN
    }.
    \textbf{a},
    Distribution of entropy, maximum softmax probability (MSP), and softmax entropy (SE) for correctly classified, misclassified, and out-of-distribution inputs.
    \textbf{b},
    The separability of three indicators on the correctly classified, incorrectly classified and out-of-distribution (OoD) samples.
    The negative value of separability implies that the MSP and SE incorrectly indicate higher uncertainty for misclassified samples compared to out-of-distribution samples.
    \textbf{c},  
    The test accuracy of different models under various strengths $\epsilon$ of FGSM adversarial attacks.
    \textbf{d}, Box plot of the entropy of the model prediction for samples in the test set, for the MNN with $\sigma_1,\sigma_2>0$ under various strengths $\epsilon$ of FGSM attacks. Orange line: median; box: upper and lower quartile.
    } 
    \label{fig:ood_attack}
\end{figure}
Another important aspect for uncertainty representation is detecting out-of-distribution samples, which refers to data of a type different from that a model is trained on. To investigate whether MNNs can reasonably express uncertainty of out-of-distribution samples, we analyze the above MNN trained on MNIST using out-of-distribution samples from the notMNIST dataset~\cite{bulatov2011notmnist}. Specifically, we compare the uncertainty expressed by output covariance (as measured by entropy) and that expressed by output mean (as measured by maximum softmax probability and softmax entropy, two common out-of-distribution indicators~\cite{hendrycksbaseline}). 
As shown in Fig.~\ref{fig:ood_attack}\textbf{a},
the distribution of covariance-based indicator (entropy) on the correctly classified, misclassified and out-of-distribution samples are better separated than that of the mean-based indicators. Additionally, the separabilities of the covariance-based indicator are significantly higher than that of the mean-based indicators (Fig.~\ref{fig:ood_attack}\textbf{b}). These results show that the entropy is more effective at separating the correctly classified, misclassified and out-of-distribution samples compared to the other two indicators, implying that the output covariance is a more appropriate uncertainty indicator than the those derived from the output mean.
Furthermore, it sheds light on why mean-based indicators can offer information about uncertainty: since the mean and covariance are non-linearly coupled, they share pertinent information regarding uncertainty.

We investigate the potential of MNNs to defend against gradient-based adversarial attacks~\cite{szegedy2013intriguing,goodfellow2015explaining}, which can significantly degrade the performance of a classifier by modifying input images imperceptible by human. We train an Heaviside MNN on the MNIST dataset and then apply fast gradient sign method (FGSM)~\cite{goodfellow2015explaining} to generate adversarial samples. 
We find that the MNN is susceptible to adversarial attacks (Fig.~\ref{fig:ood_attack}\textbf{c}). Nevertheless it indicates high uncertainty on the attacked samples by exhibiting an increase in the output entropy (Fig.~\ref{fig:ood_attack}\textbf{d}).

Here, we propose a simple modification to the network to combat gradient-based adversarial attacks. During inference, we set $\sigma_1$ and $\sigma_2$ in Eq.\eqref{eq:mnn_feedforward} to zero, making the Heaviside MA in the first layer (Eq.~\eqref{eq:heav_mapping}) reduce to the Heaviside function with zero derivatives. This results in significant boost in robustness in exchange of only a marginal drop in test accuracy (from $0.9797$ to $0.9745$). As shown in Fig.~\ref{fig:ood_attack}\textbf{c}, the test accuracy remains unchanged regardless of the attack strength applied, implying that the FGSM attack ineffective on this MNN. This is because FGSM and similar attack methods utilize the error gradient, which becomes zero due to the Heaviside function at the first layer, making it difficult to generate effective adversarial samples.

\subsection{{Comparison with other uncertainty quantification approaches}}

We further compare our MNN with other representative methods for quantifying uncertainty. We train MNNs on UCI regression dataset~\cite{asuncion2007uci} with mean squared error (MSE) as the training loss.
We report the log-likelihood which evaluates both the prediction (mean) and uncertainty quantification (covariance) of the model. As shown in Tab.~\ref{tab:regression_main}, our MNNs demonstrate comparable performance compared to other methods, and in some cases, even outperform them on several datasets.
Importantly, although our learning algorithm (SMUC) and loss (MSE) do not incorporate the covariance, it still effectively maximizes log-likelihood. These results further verify that the emerging covariance can faithfully capture the prediction uncertainty without supervision.

\begin{table}[h]
\caption{\textbf{Comparison of MNN and other uncertainty quantification methods}. Log-likelihood of different models trained on the UCI regression datasets over random train/test splits. The data shows the mean (standard deviation) over 20 runs.
MNN (R) represents ReLU MNN and MNN (H) represents Heaviside MNN. 
BP, probabilistic backpropagation~\cite{hernandez2015probabilistic}; 
MNVI, multiplicative noise variational inference~\cite{schmitt2022sampling}; DVI, deterministic variational inference~\cite{wu2018deterministic}; Monte Carlo (MC) dropout~\cite{gal2016dropout}; deep ensembles~\cite{lakshminarayanan2017simple}.
Bold font indicates the highest log-likelihood value for each dataset.
}\label{tab:regression_main}
\setlength{\tabcolsep}{.75mm}{
\begin{tabular}{lcclccccc}
\toprule[1.5pt]
& boston    & concrete    & \multicolumn{1}{c}{kin8}  & energy  & power  & protein & wine & yacht   \\  \midrule
PBP         & -2.57 (0.09)  & -3.16 (0.02)  & 0.90 (0.01)  &-2.04 (0.02) & -2.84 (0.01) & -2.97 (0.00) & -0.97 (0.01) & -1.63 (0.02) \\
MNVI  & -2.43 (0.02) & -3.05 (0.01) & 1.15 (0.01)  & -1.33 (0.05) & {\bf -2.66 (0.01)}  &   -2.99 (0.01)    & -0.96 (0.01)  & -0.37 (0.02) \\
DVI  & {\bf -2.41 (0.02)} & -3.06 (0.02) & 1.13 (0.00)   & {\bf -1.01 (0.06)} & -2.80 (0.00)  & -2.85 (0.01)          &  -0.90 (0.02)   & -0.47 (0.03)    \\
Ensemble & -2.41 (0.25) & -3.06 (0.18)         & {\bf 1.20 (0.02)}   & -1.38 (0.22) & -2.79 (0.00)          & -2.83 (0.02)          & -0.94 (0.12)         & -1.18 (0.21)          \\
MC Dropout                          & -2.46 (0.25) & {\bf -3.04 (0.09)}         & 0.95 (0.03)                      & -1.99 (0.09) & -2.89 (0.01)          & -2.80 (0.05)          & -0.93 (0.05)         & -1.55 (0.12)          \\
\midrule
MNN (R)  & -2.44 (0.00)   &	-3.59 (0.01)  &	0.92 (0.00) & -2.30 (0.01)	&-2.92 (0.00) &	-0.97 (0.00) &{\bf -0.75 (0.00)}	& {\bf -0.29 (0.06)}     \\
MNN (H)  & -2.65 (0.03)   & -3.28 (0.01)  & 0.71 (0.00)	& -1.31 (0.01)	&-2.94 (0.00) &{\bf -0.92 (0.00)} & -0.90 (0.01)	&-0.40 (0.00)      \\\bottomrule[1.5pt]
\end{tabular}
}
\end{table}

\section{{Further analysis}}
\subsection{Mechanism of uncertainty representation by emerging covariance}
\begin{figure}[h]
    \centering
    \includegraphics[width=\textwidth]{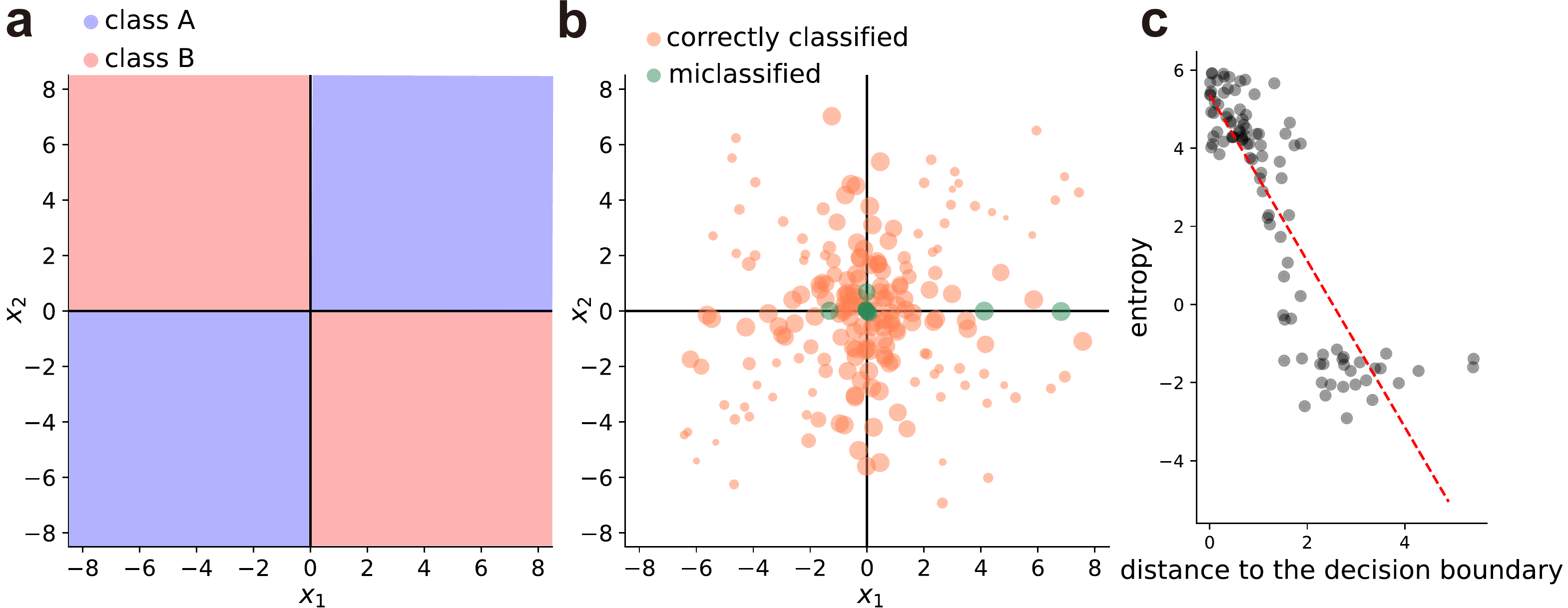}
    \caption{\textbf{Mechanism of uncertainty representation by emerging covariance.} 
    \textbf{a},
    Illustration of a binary classification dataset, where a sample in $\mathbb{R}^2$ belongs to class A (or class B) if the sign of its two coordinates $x_1, x_2$ are the same (or different). Solid lines indicate the class boundary of the dataset. %
    \textbf{b},
    Scatter plot of training samples randomly drawn from a Gaussian distribution. The misclassifed samples are typically located close to the decision boundary. The size of the bubbles indicates the magnitude of output entropy. 
    \textbf{c},
    Negative correlation is found between the distance to the decision boundary and the output entropy, with a coefficient of determination of 0.7196. 
    }
    \label{fig:interpert}
\end{figure}

To better understand the mechanism underlying uncertainty representation by emerging covariance, we train an MNN for a binary classification task on $\mathbb{R}^2$. Since the input has a spatial embedding in $\mathbb{R}^2$, this task allows us to directly analyze the relationship between the uncertainty conveyed by the MNN about each input sample and their spatial properties. 
As shown in Fig.~\ref{fig:interpert}\textbf{a}, the trained MNN learns the correct class for most input samples (with an accuracy of $0.9948$). Moreover, misclassified samples are typically located close to the decision boundary of the model (Fig.~\ref{fig:interpert}\textbf{b}) and have larger entropy. 
Figure.~\ref{fig:interpert}\textbf{c} shows that the average entropy decreases with samples' distance to the decision boundary with a correlation coefficient of $-0.85$, hence the closer a sample is to the decision boundary, the higher the output entropy. 
This explains why the adversarial attacks increase the entropy associated with each sample (Fig.~\ref{fig:ood_attack}\textbf{d}): they push the samples toward the decision boundary of the model~\cite{cao2017mitigating}, thus increasing the entropy of model prediction.

This observation leads to the question why samples closer to the decision boundary have higher entropy. This can be understood conceptually by considering the stochastic network (Eq.~\eqref{eq:stoc_network}) underlying the MNN. The stochastic state at each layer serves as input to the remaining part of the network, which acts as a classifier. When the input state is near the decision boundary of this classifier, even small fluctuations can cause substantial output variations. This leads to significant input-output sensitivity which causes the network to amplify the stochastic fluctuation of its input, resulting in high entropy. To provide a clearer theoretical understanding, 
we derive an explicit expression for the entropy and the input's distance to the decision boundary for a binary logistic regression problem (\mm).

Above insight allows us to formulate the following explanation for the high entropy of model prediction on the out-of-distribution samples. Consider the scenario where an MNN is trained for an arbitrary classification problem. 
Initially, the decision boundaries of the model are randomly distributed throughout the entire input space. 
As learning progresses, the model adapts the decision boundaries to fit the actual class boundary of the dataset of in-distribution samples. 
Since the entropy is lower when they are far away from the decision boundaries (as we have shown in Fig.~\ref{fig:interpert}\textbf{b}), only the small set of samples near the class boundary have high entropy.
In the out-of-distribution region, however, the decision boundaries lack a structured distribution and are randomly scattered throughout the region, hence increasing the chance that an out-of-distribution sample is close to one of the decision boundaries.
As a result, the entropy of out-of-distribution samples tends to be higher compared to that of in-distribution samples.

\subsection{\ma{Effect of the covariance}}
\begin{figure}[h]
    \centering
    \includegraphics[width=0.7\textwidth]{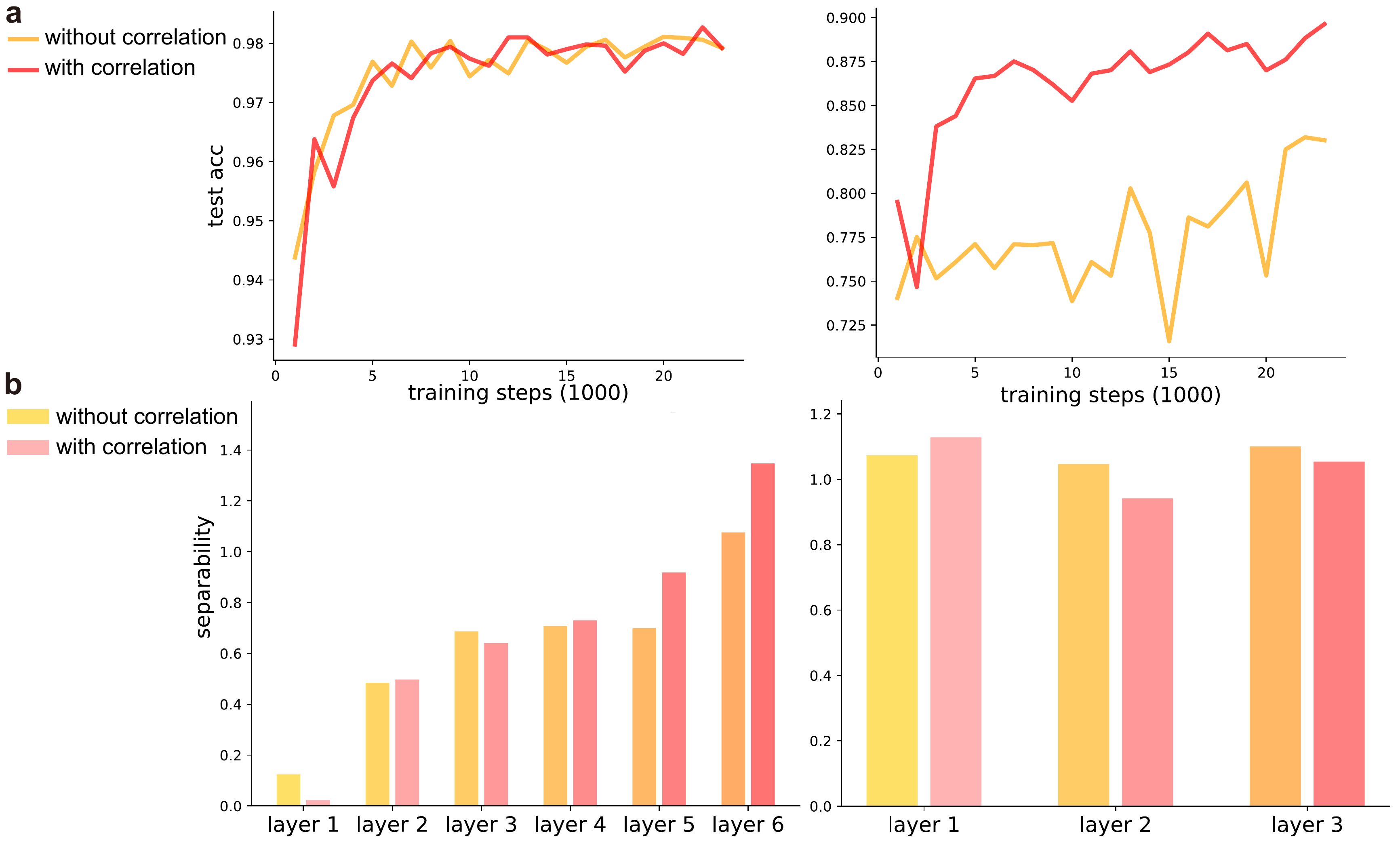}
    \caption{\textbf{Comparison of MNNs with and without correlation.} 
    \textbf{a,} Test accuracy across training procedure for MNNs with and without correlation on MNIST (left) and CIFAR-10 (right) classification.
    \textbf{b,} Entropy separability across layers procedure for MNNs with and without correlation on MNIST (left) and CIFAR-10 (right) classification. 
    }
    \label{fig:cov_vs_no_cov_all}
\end{figure}

We further investigate the effect of the correlation between neurons on model learning and uncertainty quantification. 
We repeat the experiments presented in Fig.~\ref{fig:classifier_all}, while manually setting all off-diagonal elements of the covariance matrices to zero during the training and inference of MNNs.
As shown in the left panel of Fig.~\ref{fig:cov_vs_no_cov_all}\textbf{a}, for MNIST classification (pure MNN), the MNN without correlation has similar performance in terms of test accuracy, However, for CIFAR-10 classification (mixed MNN), MNN without correlation learns much slower compared to the one with correlations, as shown in the right panel of  Fig.~\ref{fig:cov_vs_no_cov_all}\textbf{a}.
In terms of uncertainty quantification, we compare the separability of the entropy on the correctly classified and misclassified samples across layers for MNIST  the left panel of (Fig.~\ref{fig:cov_vs_no_cov_all}\textbf{b}) and CIFAR-10 the right panel of (Fig.~\ref{fig:cov_vs_no_cov_all}\textbf{b}) classification. 
It is observed that for CIFAR-10, there is only a slight decrease in separability when omitting the correlation between neurons. On the other hand, for MNIST, there is a relatively larger decrease in separability. This difference can be attributed to the fact that in the mixed MNN, the inputs are initially decorrelated by the convolution layers. As a result, the correlations in the layers that utilize MAs become less crucial for uncertainty quantification. However, as demonstrated in Figure~\ref{fig:cov_vs_no_cov_all}b, these correlations still play certain role in facilitating faster learning.

\section{{Discussion}}
\subsection{{Sufficiency of the second-order cumulant approximation}}
In this study, we have approximated a stochastic network up to its second-order cumulant while omitting the higher-order cumulants. The validity of this approximation is supported by both empirical evidence and theoretical analysis.
In ref.~\cite{schneidman2006weak}, the authors analyzed neural data of vertebrate retina and discovered that the pairwise correlations of biological networks can  accurately predict the collective effects of the system without considering higher-order interactions. 
In ref.~\cite{dahmen2016correlated}, the authors discovered that in a weakly correlated state, the contributions of higher-order cumulants on network dynamics are significantly smaller, implying that the first two cumulants provide a reasonably accurate representation of the network dynamics.

\subsection{{Relation to different learning schemes}}
Several methods for uncertainty quantification represent the stochasticity of neural networks in a deterministic fashion~\cite{hernandez2015probabilistic,haussmann2020sampling,wu2018deterministic,schmitt2022sampling}.
Our approach differs from their methods in that the neural network is trained with the SMUC learning rule, 
in which only the mean but not the covariance is supervised, while theirs fall under the BVI formulation using the SMSC learning rule, in which both mean and covariance are supervised during training, as illustrated in \mm.
The MNN trained with SMUC can demonstrate superior performance on uncertainty quantification (Tab.~\ref{tab:regression_main}).
Although the present study has focused on probabilistic learning under the supervised setting, future works can future explore this idea for unsupervised learning, where labels are unavailable.
For example, Barlow Twins method~\cite{zbontar2021barlow} trains an encoder using a covariance-based loss function to reduce redundancy. This suggests it is possible to train MNNs using a unsupervised mean supervised covariance (UMSC) approach for unsupervised learning tasks.

\subsection{{Inspiration for learning in the brain}}
Our results leads to an intriguing prediction about the learning rules employed by the brain.
Although biological neural systems is inherently probabilistic due to the presence of various sources of noise and variability~\cite{averbeck2006neural}, we expect that learning of probabilistic computation can be realized by adjusting synaptic connectivity through gradient descent via the mean firing rate of neurons.
Our prediction is founded on two key observations: 
first, we have demonstrated that the MNN derived from the leaky integrate-and-fire neuron model, which is commonly used for modeling spiking neuronal activity in the brain, can capture prediction uncertainty through its emerging covariance (Fig.~\ref{fig:classifier_lif_all});
second, it is commonly hypothesized that the brain adjusts synaptic connectivity in a manner similar to gradient descent~\cite{goodfellow2016deep}.
To further test our predictions, we propose investigating whether neural dynamics observed in the brain can be replicated by MNNs.

\section{Conclusion}
Our research introduces two perspectives on probabilistic computation. 
First, we demonstrate that complex probabilistic computation can be conducted by combinations of simple nonlinearities that captures truncated transition probabilities up to their second-order cumulants.
Second, we reveal that output covariance spontaneously emerging from the nonlinear coupling between the mean and covariance of network states faithfully captures the prediction uncertainty.
Our probabilistic computation framework acts as a natural extension to conventional ANNs, seamlessly incorporating uncertainty quantification into learning and inference. Our methodology demonstrates the feasibility of probabilistic computation within the current deterministic setting, paving the way for its integration into large-scale AI systems.

We derive a set of moment activations (MAs, Eq.~\eqref{eq:mean_map}-\eqref{eq:cov_map}) that propagate nonlinearly coupled cumulants and enable gradient-based learning of probabilistic computation.
We then utilize the MAs to construct a new class of neural network called the moment neural network (MNN, Eq.~\eqref{eq:mnn_feedforward}) which approximates a parameterized stochastic neural network (Eq.~\eqref{eq:stoc_network}) up to its second cumulants.
Unlike conventional ANNs that primarily focus on nonlinear transformations of the mean, MNNs also incorporate nonlinear coupling between the mean and covariance.
This capability enables direct training through gradient descent, eliminating the need for MC sampling methods.
Moreover, MNNs offer a unique advantage in analyzing the dynamics of stochastic networks by explicitly expressing the nonlinear coupling between the mean and covariance, which is implicit in MC methods.

Next, we develop a learning algorithm called SMUC for training MNNs (Eq.~\eqref{eq:mnn_feedforward}-\eqref{eq: update_theta}), where the loss is backpropagated only through the mean while treating the covariance as constants during gradient descent. Through extensive experiments (Fig.~\ref{fig:classifier_all}-\ref{fig:ood_attack}, Tab.~\ref{tab:regression_main}), we demonstrate that the covariance spontaneously emerging from its nonlinear coupling with the mean effectively captures the prediction uncertainty.
This challenges the conventional wisdom that it is necessary to supervise both the prediction and the prediction uncertainty for uncertainty quantification.

\section*{Code Availability}
The code for training the moment neural network (MNN) through supervised mean unsupervised covariance (SMUC) is available without restrictions on GitHub (https://github.com/AwakerMhy/probabilistic-computing-mnn).

\section*{Acknowledgments}
Supported by STI2030-Major Projects (No. 2021ZD0200204); supported by Shanghai Municipal Science and Technology Major Project (No. 2018SHZDZX01), ZJ Lab, and Shanghai Center for Brain Science and Brain-Inspired Technology; supported by the 111 Project (No. B18015).

\section*{Competing interests}
The authors declare no competing interests.

\section*{Materials \& Correspondence}
Correspondence and requests for materials should be addressed to J.F.

\section*{Appendix}
\subsection*{Moment neural network (MNN)}
In general, it is hard to solve $\mathbf{v}^{(l)}$ in the stochastic neural network Eq.~\eqref{eq:stoc_network} analytically. 
Nevertheless, under steady-state distribution, we can approximate $\mathbf{v}^{(l)}$ by a Gaussian distribution $\mathcal{N}(\bar{\bm{\mu}}^{(l)},C^{(l)})$ using mean field approximation (\si). In this way, the signals received by the neurons at $l$-th layer obey a Gaussian distribution $\mathcal{N}(\bar{\bm{\mu}}^{(l)},\bar{C}^{(l)})$, with $\bar{\bm{\mu}}^{(l)}=W^{(l-1)}\bm{\mu}^{(l-1)}+\mathbf{b}^{(l-1)}$ and $\bar{C}^{(l)}=W^{(l-1)}C^{(l-1)}(W^{(l-1)})^{\mathrm{T}}+\sigma_l^2I$.
Based on the linear response theory~\cite{de2007correlation} (\si), we obtain a set of nonlinear mappings dubbed {\em moment activations (MAs)} describing how the neurons transfer the first two cumulants of received signals to that of the output signal.
See \si~for the derivation of the MAs (Eq.~\eqref{eq:mean_si}, \eqref{eq:var_si}, and \eqref{eq:cov_si}).

\subsection*{\ma{LIF MNN}}
In~\cite{feng2006dynamics,lu2010gaussian}, the authors analyzed the stochastic dynamics of a spiking neuronal network (SNN) by a set of nonlinear mapping describing how the neurons transfer the input spike signals from other neurons. 
Concretely, the authors studied a network of $N$ neurons with the dynamics of each neuron applying the leaky integrate-and-fire (LIF) neuron model
\begin{align}\label{eq:LIF}
    \dfrac{dV_i}{dt} = -LV_i+\sum_{j=1}^Nw_{ij}S_j,\quad i=1,\ldots,N,
\end{align}
where $V_i$ is the membrane potential of the $i$-th neuron, $L$ is the leaky conductance, $w_{ij}$ is the synaptic weight from neuron $j$ to neuron $i$, and $S_j$ is the spike train from pre-synaptic neurons. 
When $V_i$ reaches the firing threshold $V_{\text{th}}$
the neuron will release a spike which is transmitted to other connected neurons, and then $V_i$ is reset to the resting potential $V_{\text{res}}$ and enters a refractory period $T_{\text{ref}}$.
The corresponding moment activations, which we refer to as {\em LIF moment activations (LIF MAs)}, are as follows~\cite{lu2010gaussian}
\begin{align}
    &m_{v}(\bar{\mu},\bar{C}) =  \big(T_{\rm ref} + \frac{2}{L}\int_{I_{\rm lb}}^{I_{\rm ub}} g(x) dx\big)^{-1},\label{eq:mu_lif}\\
    &C_{v}(\bar{\mu},\bar{C}) = \frac{8}{L^2}m_{v}(\bar{\mu},\bar{C})^3\textstyle\int_{I_{\rm lb}}^{I_{\rm ub}} h(x) dx,\label{eq:sigma_lif}\\
    &\chi(\bar{\mu},\bar{C}) = \frac{\partial}{\partial\bar{\mu}}  m_{v}(\bar{\mu},\bar{C})\label{eq:cov_lif}
\end{align}
where $T_{\rm ref}$ is the refractory period with integration bounds $I_{\rm ub}(\bar{\mu},\bar{C}) = \tfrac{V_{\rm th}L-\bar{\mu}}{\sqrt{L\bar{C}}}$ and
$I_{\rm lb}(\bar{\mu},\bar{C}) = \tfrac{V_{\rm res}L-\bar{\mu}}{\sqrt{L\bar{C}}}$.
The constant $L$, $V_{\rm res}$, and $V_{\rm th}$ are identical to those in Eq.~\eqref{eq:LIF}.
The pair of Dawson-like functions $g(x)$ and $h(x)$ appearing in Eq.~\eqref{eq:mu_lif} and Eq.~\eqref{eq:sigma_lif} are
\begin{align}
    g(x)=e^{x^2}\int_{-\infty}^x e^{-u^2}du,\quad h(x)=e^{x^2}\int_{-\infty}^x e^{-u^2}[g(u)]^2du.
\end{align}
It is important to highlight that in the LIF MAs (Eq.~\eqref{eq:mu_lif}-\eqref{eq:cov_lif}), the mean is the average number of the spike in a unit time interval, the variance is the variance of the number of the spike in a unit time interval, and the covariance between two neuron $i$ and $j$ is the multiplication of their variances and the coherence of their spike train~\cite{de2007correlation}.
Using the LIF MAs, we can construct the corresponding MNN dubbed {\em LIF MNN} with learnable weight matrix. After training, the MNN can be reverted into the corresponding SNN~\cite{qi2023spike}, similar as the ReLU and Heaviside MNNs. 
In comparison to the ReLU and Heaviside MAs, the computation of LIF MAs involves higher numerical complexity due to the Dawson-like functions, which can present challenges in implementing them for large-scale models. Nonetheless, the LIF MNN exhibits well-defined biological plausibility, making it a valuable tool for analyzing the neural circuit mechanisms underlying brain coding and learning processes~\cite{ma2022dynamics}.
For implementation, we set $V_{\rm th}=20$ mV, $V_{\rm res}=0$ mV, $T_{\rm ref}=5$ ms, and $\tau=1/L=20$ ms, following~\cite{lu2010gaussian}.
and conduct the LIF MAs by the efficient numerical algorithm developed in~\cite{qimnn2022}.

\subsection*{Metrics for quantifying uncertainty}
{\bf Entropy. }
The output of the MNN can be considered as a multi-dimensional Gaussian distribution $\mathcal{N}(\bm{\mu}_{y},C_y)$, whose entropy is calculated as
\begin{align}\label{eq:entropy}
        H\big(\mathcal{N}(\bm{\mu}_{y},C_y)\big) = \frac{n}{2}(1+\log 2\pi) + \frac{1}{2}\log \mathrm{det}[C_y],
\end{align}
where $n$ is the dimension of the output. If $C_y$ is singular, we have $\mathrm{det}[C_y]=0$, which leads to entropy of infinite negative value. In such scenarios, we compute the entropy in the linear subspace of $\mathbb{R}^n$ where the covariance matrix $C_y$ is full-rank through SVD and use the dimension of this subspace as the effective dimension for calculation.
It is important to note that the entropy indicators do not incorporate the output mean $\bm{\mu}_{y}$, which is supervised during training. Instead, this indicator solely rely on the output covariance $C_y$, which emerges from the nonlinear coupling with the mean.
Additionally, we can calculate the entropy at each layer of the MNN. Given the mean and covariance of the $l$-th layer as $\bm{\mu}^{(l)},C^{(l)}$, the entropy at the $l$-th layer is
\begin{align}
        H\big(\mathcal{N}(\bm{\mu}_v^{(l)},C^{(l)})\big) = \frac{m^{(l)}}{2}(1+\log2\pi) + \frac{1}{2}\log \mathrm{det}[C^{(l)}].
\end{align}
\noindent{\bf Maximum softmax probability (MSP)}
Given the output mean $\bm{\mu}_{y}$, we can define a probability distribution using softmax function
\begin{align}\label{eq:softmax_p}
    p_i = \frac{\exp\{\mu_{y,i}\}}{\sum_{j=1}^n\exp\{\mu_{y,j}\}},\quad i=1,\ldots,n.
\end{align}
The maximum softmax probability (MSP) has been widely employed as an indicator for detecting out-of-distribution samples~\cite{hendrycksbaseline}. The MSP is determined by calculating the maximum probability, denoted as $\max_{i} p_i$, from the softmax output.
Lower values of MSP indicate higher uncertainty in the model's predictions. 

\noindent{\bf Softmax entropy (SE).}
Another indicator utilized is the entropy of the distribution ${p_i}$, as proposed by ~\cite{steinhardt2016unsupervised}. The entropy, referred to as softmax entropy (SE) in this context, is computed as $-\sum_{i=1}^n p_i \log p_i$. It is important to distinguish this entropy from the entropy calculated by MNNs.
The SE provides a measure of the uncertainty in the softmax output distribution. Higher SE values indicate greater uncertainty in the model's predictions.

\subsection*{Separability}
Suppose we aim to quantify the degree of separation for a variable between two populations, labeled as 1 and 2. In such cases, we can calculate the separability as
\begin{align}
    \frac{\mu_1-\mu_2}{\sqrt{\sigma_1^2+\sigma_2^2}},
\end{align}
where $\mu_1$ and $\mu_2$ are the means of the variable in population 1 and 2 respectively, and $\sigma_1^2$ and $\sigma_2^2$ are variances of the variable in population 1 and 2 respectively.
A higher absolute value of the separability indicates a greater statistical distinction between the two populations.
When calculating the separability of entropy between two sample populations in the main paper, we designate population 1 as the population we expect to have a higher entropy (misclassified samples, for instance) compared to the other population (correctly classified samples, for instance).

\subsection*{Mean square error (MSE) and log likelihood (LL)}
Supposed that there are $N_{\mathrm{test}}$ samples in the test set, and for the $i$-th sample in the test set, the model given the output mean $\bm{\mu}_{y,i}\in\mathbb{R}^d$, covariance $C_{y,i}\in\mathbb{R}^{d\times d}$ and the corresponding ground-truth output is $\mathbf{y}_i\in\mathbb{R}^d$. The mean square error (MSE) measurement is calculated as
\begin{align}
    \mathrm{MSE} = \frac{1}{N_{\mathrm{test}}} \sum_{i=1}^{N_{\mathrm{test}}}\left\|\mathbf{y}_i-\bm{\mu}_{y,i}\right\|^2,
\end{align}
which evaluates the accuracy of the model prediction while does not include the uncertainty quantification ability of the model.
The log likelihood (LL) measurement is calculated as
\begin{align}
    \mathrm{LL} = 
    -\frac{1}{N_{\mathrm{test}}} \sum_{i=1}^{N_{\mathrm{test}}} \frac{1}{2}\big(\log\mathrm{det}[ 2\pi C_{y,i}] +(\mathbf{y}_i-\bm{\mu}_{y,i})^{\mathrm{T}}C_{y,i}^{-1}(\mathbf{y}_i-\bm{\mu}_{y,i}) \big ),
\end{align}
which evaluates both the prediction (mean) and uncertainty quantification (covariance) of the model. 

\subsection*{Training implementations}
By default, we utilize the cross-entropy loss function as the objective function and employ the Adam optimizer~\cite{kingma2014adam} for training, without any additional specifications. Additionally, we set $\sigma_l$ to a value of $0.2$ for all layers. 

\noindent{\bf Classification on MNIST~\cite{krizhevsky2009learning}. }
We train ReLU and Heaviside MNNs respectively, whose architecture consisting of six fully connected layers with dimensions of $784, 392, 196, 96, 48, 24, 10$, respectively.
We set the training batch size as $128$, the learning rate as $0.0005$, weight decay factor as $0.001$ and the training epochs as $50$.
For LIF MNN, we construct a fully connected LIF MNN on MNIST dataset for classification, where the dimension at each layer are 784, 392, 96 and 10, and we add a batch-normalization layer before each nonlinearity, following~\cite{qi2023spike}. We apply Adam~\cite{kingma2014adam} as the optimizer. The training batch-size is 128, and epoch is 15, the learning rate is 0.0005, and the factor of weight decay is 0.0001.

\noindent{\bf Classification on CIFAR-10~\cite{krizhevsky2009learning}. }
We train a mixed MNN with ReLU and Heaviside MAs respectively. The network consists of 10 layers of convolutional layer and three fully connected MA layers. The construction of these convolutional layers follows the design of VGG13~\cite{simonyan2014very}.
The dimensions of the MA layers are set as $64, 32, 10$, respectively.
We set the training batch size as $128$, the learning rate as $0.0005$, weight decay factor as $0.001$ and the training epochs as $120$, using random crop and random horizontal flip as data augmentation.

\noindent{\bf Out-of-distribution detection. }
We train a fully connected ReLU MNN on the MNIST dataset, and the dimension of each layer is $784, 392, 196, 96, 48, 24, 10$. The training batch-size is 128, and epoch is 50, the learning rate is 0.0005, and the factor of weight decay is 0.0001.
After training, we use the notMNIST~\cite{bulatov2011notmnist} dataset as the out-of-distribution samples. NotMNIST dataset consists of $28\times28$ grayscale images of alphabets from $A$ to $J$.

\noindent{\bf Adversarial attack defense and awareness. }
We train an fully connected Heaviside MNN on MNIST, where the dimension of each layer is $784, 392, 196, 96, 48, 24, 10$. The training batch-size is 128, and epoch is 50, the learning rate is 0.0005, and the factor of weight decay is 0.0001.
We apply the fast gradient sign method (FGSM)~\cite{goodfellow2015explaining} for generating adversarial samples on the test set. FGSM adds an optimal max-norm constrained perturbation to the input $\mathbf{x}$, and this perturbation is calculated as $\epsilon\mathrm{sgn}(\bigtriangledown_{\mathbf{x}}\mathcal{L}(\theta,\mathbf{x},\mathbf{y}))$,
where $\mathbf{x},\mathbf{y}$ is an input-output pair from dataset, $\theta$ is the parameters of the trained model, and $\epsilon>0$ controls the strength of the perturbation. 
The FGSM finds the direction along which the model output changes the most and adds small perturbation along that direction.

\noindent{\bf Regression problem}
We train both ReLU MNN and Heaviside MNN through SMUC for reducing the mean square error (MSE) loss on the predicted outputs and ground-truth outputs. We normalize the dimensions of all the datasets so that their mean are $0$ and standard deviation are $1$. The MNN is fully connected and has one hidden layer of width $50$, which is the same as that in~\cite{schmitt2022sampling}.
We utilize the Adam optimizer with a weight decay factor of zero for optimization. Across all datasets, the training batch size is fixed at 128, the learning rate is set to 0.001, and the number of epochs is set to 500, except for the power and protein datasets, where we set the epoch to 20. We assign a value of 0 to $\sigma_2$ and choose the best results from among the options for $\sigma_1$ which include 0.02, 0.05, and 0.1.

\noindent{\bf Binary classification problem. }
We assign the categories of samples in the training dataset as follows. Given a sample $\mathbf{x}\in\mathbb{R}^2$, it belongs to class A if the sign of two coordinates $\mathbf{x}_1,\mathbf{x}_2$ are the same, otherwise it belongs to class B. 
We train a fully connected ReLU MNN. The dimensions of each layer in the network are $2, 64, 32, 16, 8, 2$, respectively, and $\sigma_l$ are all set as $0.5$. 
We generate the training set of 10,000 samples by randomly sampling from a standard Gaussian distribution on $\mathbb{R}^2$. The training batch is 128, the epoch is 20, the learning rate is 0.0005, and the factor of the weight decay is 0.00001.
Since the model fits well with the dataset (Fig.~\ref{fig:interpert}\textbf{b}), we consider that the decision boundary learned by the model aligns with the classification boundary of the dataset ($\mathbf{x}_1$ and $\mathbf{x}_2$ axes).
The distance from a point $\mathbf{x}$ to the decision boundary ($\mathbf{x}_1$ and $\mathbf{x}_2$ axes) is $d(\mathbf{x}) = \min(\left|\mathbf{x}_1\right|,\left|\mathbf{x}_2\right|)$.

\section*{\ma{Numerical verification of covariance activations and moment neural networks}}

\begin{figure}[h]
    \centering
    \includegraphics[width=\textwidth]{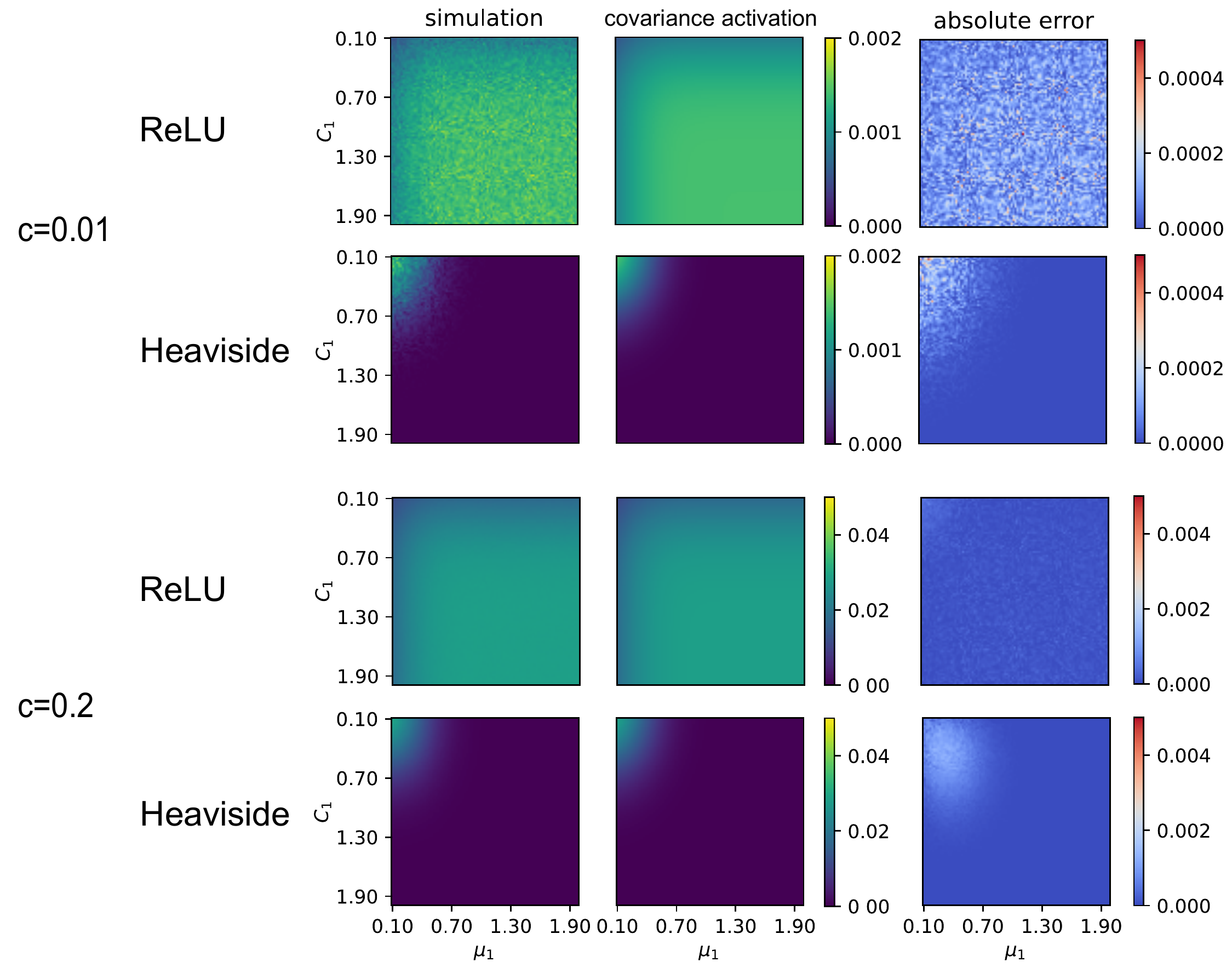}
    \caption{\textbf{Numerical verification for covariance activations.} The covariance of the signals produced by two neurons calculated using Heaviside or ReLU covariance activations is consistent with that of Monte Carlo simulations. 
    We present the results with two different correlation levels: a small correlation value of $c=0.01$ (top) and a relatively larger correlation value of $c=0.2$ (bottom).
    }
    \label{fig:cov_error}
\end{figure}

\begin{figure}[h]
    \centering
    \includegraphics[width=0.95\textwidth]{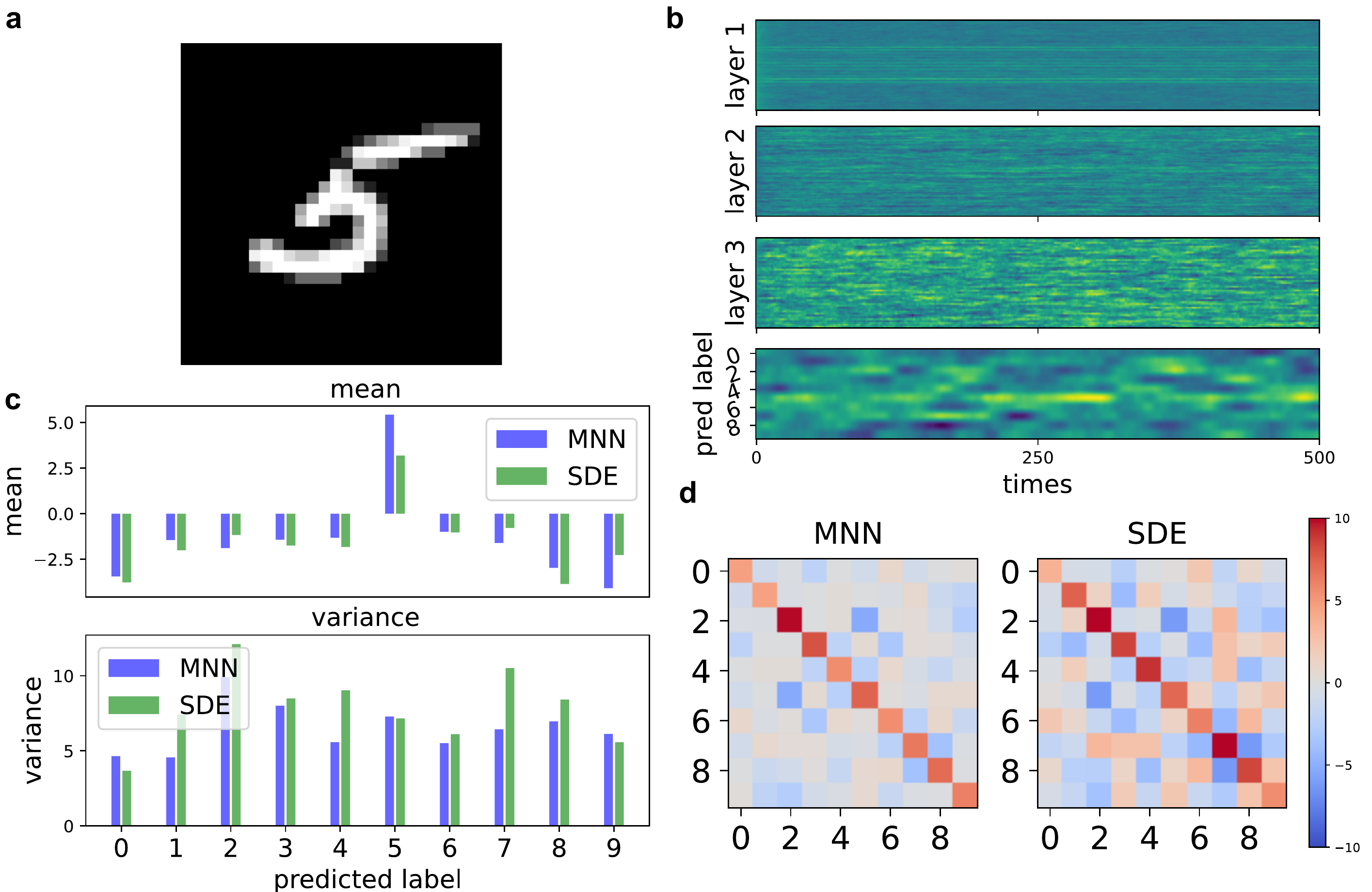}
\caption{\textbf{The Heaviside MNN (Eq.~\eqref{eq:mnn_feedforward}) approximates the corresponding stochastic network (Eq.~\eqref{eq:stoc_network}) up to its first two cumlants.}
{\bf a,} The input data from MNIST.
{\bf b,} We train an MNN on MNIST and use the parameters of the trained MNN to construct the corresponding stochastic network, and show its dynamics.
{\bf c,} The mean and variance output by MNN and estimated using Monte Carlo methods from the stochastic network. 
{\bf d,} The covariance matrix produced by MNN and estimated using Monte Carlo methods from the stochastic network. 
}
    \label{fig:verification2}
\end{figure}

We first numerically verify the accuracy of the covariance activations (Eq.~\eqref{eq:cov_si}) for ReLU and Heaviside functions.
We present the signals produced by two neurons $C_{v,ij}$ calculated using the covariance activation, and compare it with the covariance obtained through Monte Carlo simulation as shown in Fig.~\ref{fig:cov_error}. The consistency between the results obtained from MNNs and the stochastic networks validate that MNNs well approximates the underlied stochastic network up to its first two cumulants.

We then numerically verify that the ReLU and the Heaviside MNN can approximate the output mean and covariance of the corresponding stochastic neural network. 
We train a fully connected Heaviside MNN on MNIST for classification, and then use the trained parameters to construct the corresponding stochastic neural network (Eq.~\eqref{eq:stoc_network}). When provided with an input (Fig.~\ref{fig:verification2}\textbf{a}), the stochastic network generates corresponding dynamic patterns. (Fig.~\ref{fig:verification2}\textbf{b}). We estimate the output mean, variance and covariance of the stochastic network through Euler Maruyama method, and compared it with the that calculated by the Heaviside MNN (Fig.~\ref{fig:verification2}\textbf{c},\textbf{d}). 
As shown in Fig.~\ref{fig:verification2}, it is observed that the Heaviside MNN approximates the corresponding stochastic network well in terms of the output mean and covariance. 

\section*{\ma{Theoretical comparison with Bayesian variational inference framework}}
We conducted a comparison between SMUC and  Bayesian variational inference (BVI) framework. To enable a direct comparison, we rewrite the MNN using the steady-state of the stochastic network (Eq.~\eqref{eq:stoc_network}) under the mean field approximation
\begin{align}\label{eq:multi_network}
\begin{aligned}
&\mathbf{v}^{(1)} =  \mathbf{x}+\sigma^{(1)}\mathbf{z}^{(1)}\\
&\mathbf{v}^{(l)}=\mathbf{}(W^{(l-1)}\mathbf{v}^{(l-1)}+\mathbf{b}^{(l-1)}+\sigma_{l}\mathbf{z}^{(l)}),\quad l=2,\ldots,L\\
&\mathbf{y} = W^{(L)}\mathbf{v}^{(L)},
\end{aligned}
\end{align}
where $\mathbf{z}_l$ are independent standard Gaussian vectors.
After combining the bias $\mathbf{b}^{(l)}$ and the Gaussian vectors $\sigma_l\mathbf{z}^{(l)}$ at each layer, we can regard the biases as stochastic parameters, obeying $\mathcal{N}(\mathbf{b}^{(l)},\sigma_l^2I)$.
Hence, we can consider the network is parameterized by stochastic parameters $\theta$, and the distribution of $\theta$ is parameterized by the trainable parameters $\psi$ which collects all the weights $W^{(l)}$ and the bias $\mathbf{b}^{(l)}$. Denote the distribution of $\theta$ as $q_{\psi}(\theta)$.
Then the variational loss is
\begin{align}\label{eq:vi_loss}
    \mathcal{L}(\theta) = -\int q_{\psi}(\theta)\log p(\mathbf{y}|\mathbf{x},\theta)d\theta + \mathrm{KL}(q_{\psi}(\theta)||p(\theta)),
\end{align}
where $p(\theta)$ is the prior distribution of $\theta$.
The right of Eq.~\eqref{eq:vi_loss} are log-likelihood term and KL-divergence term respectively.
We approximate the prediction probability $\int p(\mathbf{y}|\mathbf{x},\theta)q_{\psi}(\theta)d\theta$ up to its second cumulants by a Gaussian distribution $\mathcal{N}(\mu(\mathbf{x},\psi),C(\mathbf{x},\psi))$, where the mean and covariance are functions of $\mathbf{x}$ and $\bm{\psi}$. 
Suppose that the loss of the task is mean square error, then the log-likelihood term of variational inference is 
\begin{align}\label{eq:mean_cov_super}
    \mathbb{E}_{\bm{\eta}\sim\mathcal{N}(\mu(\mathbf{x},\psi),C(\mathbf{x},\psi))}[\left\|\bm{\eta}-\mathbf{y}\right\|^2] =\left\|\mu(\mathbf{x},\psi)-\mathbf{y} \right\|^2 +\mathrm{Tr}(C(\mathbf{x},\psi)).
\end{align}
In our training procedure SMUC, however, only the mean is supervised for minimizing $\left\|\mu(\mathbf{x},\psi)-\mathbf{y} \right\|^2$, and the covariance spontaneously emerges via the nonlinear coupling with the mean.
Thus, we call BVI as {\em supervised mean supervised covariance} (SMSC), as it supervises both mean and covariance.

\section*{\ma{Sharing the covariances
within a mini-batch during training}}
We empirically find that the complexity for training MNNs can be further reduced without significantly sacrificing performance.
Addressing computational complexity is a common challenge in probabilistic computation algorithms, particularly when dealing with high-dimensional inputs or large datasets~\cite{abdar2021review}.
For MNN and other probabilistic computation algorithms that explicitly calculate covariance matrices such as~\cite{wu2018deterministic}, the main overhead of computation is on the calculating and storing the covariance matrices, which arises $O(d^3)$ time complexity and $O(d^2)$ space complexity.
In the case of MNNs and other probabilistic computation algorithms that involve explicit calculations of covariance matrices, such as the approach described in~\cite{wu2018deterministic}, the primary computational overhead lies in the calculation and storage of these covariance matrices. This process incurs a time complexity of $O(d^3)$ and a space complexity of $O(d^2)$, where $d$ represents the data dimensionality.
When the covariance matrices are involved into backpropagation, the computational complexity becomes even higher.
Luckily, we do not backpropagation on the covariance in MNN. Hence the main complexity comes from the feed-forward propagation.
Supposed that the batch size during training is $N_{\mathrm{batch}}$, then to compute a covariance matrix of vectors of dimension $d$ for the whole batch, the space and time complexity are $O(N_{\mathrm{batch}}d^2)$ and $O(N_{\mathrm{batch}}d^3)$ respectively.
To make MNNs more scalable, instead of calculating the covariance matrix for each input in the mini-batch during training, we calculate a single covariance matrix for each mini-batch. 
Supposed that there is a mini-batch of input $\{x_1,x_2,\ldots,x_{N_{\mathrm{batch}}}\}$, at layer $l$, the corresponding means are $\{\mu_1^{(l)},\mu_2^{(l)},\ldots,\mu_{N_{\mathrm{batch}}}^{(l)}\}$. We calculate the mini-batch wise average mean
\begin{align}
    \mu^{(l)} = \frac{1}{N_{\mathrm{batch}}}\sum_{k=1}^{N_{\mathrm{batch}}}\mu_k^{(l)},\quad l=1,2,\ldots,L
\end{align}
and use it to calculate the covariance matrix, which is shared for the inputs in this mini-batch
\begin{align}
    C^{(l)} = C_v(W^{(l-1)}\mu^{(l-1)},W^{(l-1)}C^{(l-1)}(W^{(l-1)})^{\mathrm{T}}),\quad l=2,\ldots,L,
\end{align}
where $C^{(1)}=C^{\mathrm{in}}$, which is set as the same for all the input.
In this way, the space and time complexity are reduced to $O(d^2)$ and $O(d^3)$ respectively.

\begin{figure}[h]
    \centering
    \includegraphics[width=\textwidth]{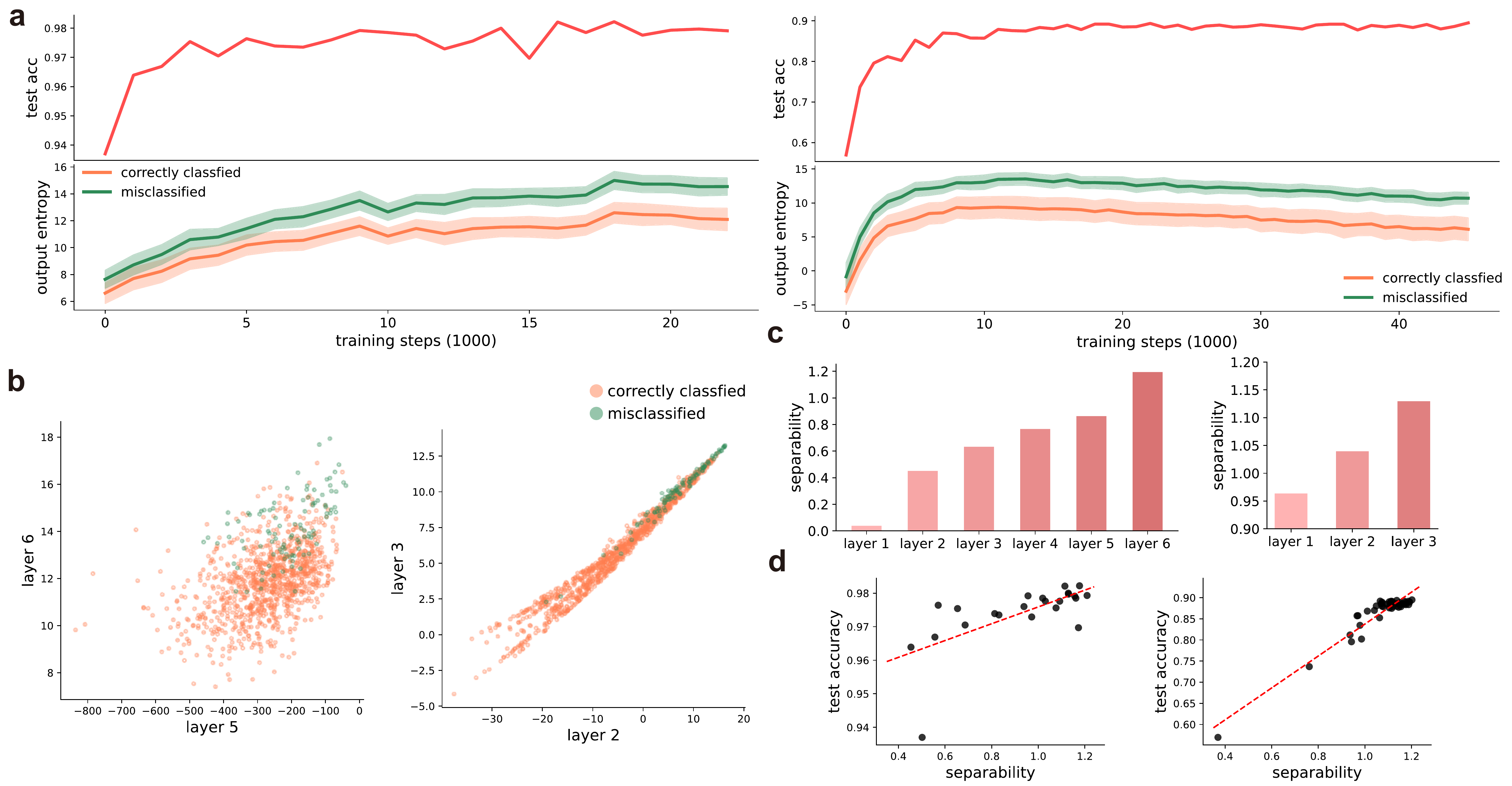}
    \caption{{\bf Emerging covariance of MNNs faithfully captures the prediction uncertainty when sharing the covariance within a mini-batch during training. }(Mixed) MNN trained on MNIST (CIFAR-10) for image classification using the batch-wise covariance trick.
    \textbf{a,}
    The training of MNNs on (left) and CIFAR-10 (right), with the accuracy (red line) and entropy (orange and green for correctly classified inputs and misclassified inputs respectively) on the test set.
    %
    % The shadows indicate the level of half standard deviations. As the training progresses, the entropy of correctly classified inputs and misclassified inputs diverges.
    \textbf{b,}
    The entropy of correctly classified and misclassified inputs in the last two layers of MNIST (left) and CIFAR-10 (right).
    %The misclassified inputs result in relatively higher entropy in both layers.
    \textbf{c,}
    The entropy separability of correctly classified and misclassified inputs on MNIST (left) and CIFAR-10 (right) increases as the layer index increases.
    \textbf{d,}
    The relationship between the separability and test accuracy during training on MNIST (left) and CIFAR-10 (right) was analyzed using linear regression. %The red dashed line represents the regression line, with 
    The coefficients of determination of 0.4334 for MNIST and 0.8994 for CIFAR-10.
    }
    \label{fig:classifier_all_batch}
\end{figure}

\begin{figure}[h]
    \centering
    \includegraphics[width=0.8\textwidth]{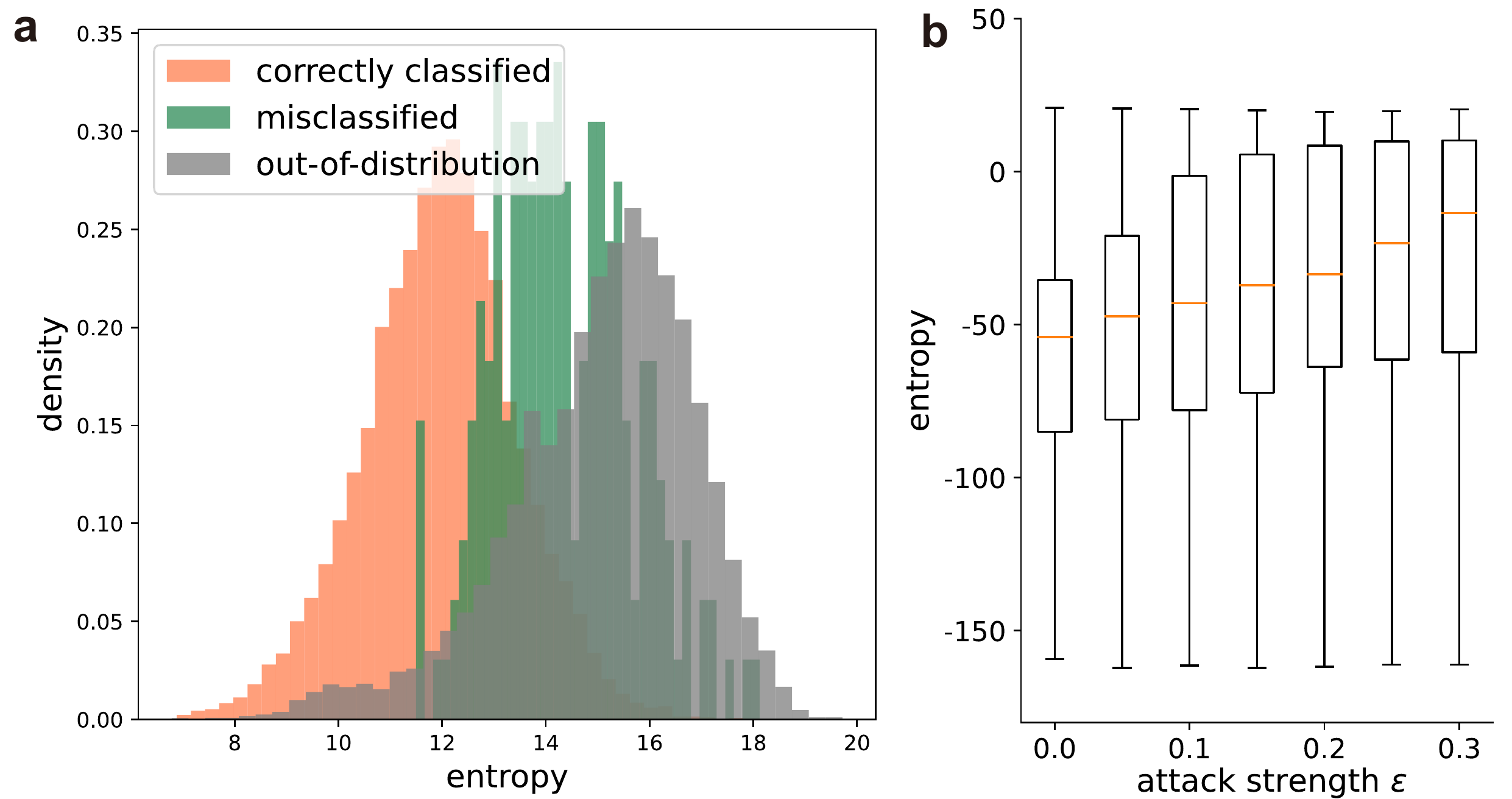}
    \caption{\textbf{The emerging covariance of MNN using the batch-wise covariance tricks for out-of-distribution detection and adversarial attack awareness.} {\bf a,} The distribution of entropy on correctly classified, misclassified and out-of-distribution input. {\bf b,} If we do not turn off the variance, although the model is vulnerable to adversarial attack, it can be aware of the attack, reflecting on the increasing of the entropy on the attacked test set.}
    \label{fig:ood_attack_batch}
\end{figure}

We report the performance of MNN when sharing the covariance in a mini-batch (\mm). As shown in Fig.~\ref{fig:classifier_all_batch} and Fig.~\ref{fig:ood_attack_batch}, the emerging covariance in the MNNs when sharing the covariance in each mini-batch also faithfully captures the prediction uncertainty, out-of-distribution detection, and adversarial attack awareness, similar as the original MNNs (Fig.~\ref{fig:classifier_all}).
We also report the corresponding performance comparison on the regression and classification tasks in Tab.~\ref{tab:regression_batch} and Tab.~\ref{tab:acc_ll_batch} respectively, and we observed that the sharing covariance within a mini-batch only causes marginal degradation compared to the original MNNs (Tab.~\ref{tab:regression} and Tab.~\ref{tab:acc_ll}).

\begin{table}[h]
\centering
\caption{{\bf Regression results of MNN when sharing the covariance within a mini-batch during training.} We report the average log-likelihood (LL) on the UCI regression datasets over random train/test splits.
MNN (H) represents Heaviside MNN, and MNN (R) represents ReLU MNN.
The data shows the mean (standard deviation) over 20
runs.}\label{tab:regression_batch}
\setlength{\tabcolsep}{.45mm}{
\begin{tabular}{lcccccccc}
\toprule[1.5pt]
                & boston      & concrete             & kin8        & energy       & power                 & protein               & wine                 & yacht                \\ \midrule
MNN (R)      & -2.46 (0.01)	 &-3.58 (0.00)	 &1.09 (0.00)	&-2.33 (0.01)	&-2.92 (0.00)	&-0.98 (0.00)	&-0.72 (0.01)	&-0.34 (0.07)    \\
MNN (H)      &-2.67 (0.04)	&-3.28 (0.02)&	0.73 (0.00)	&-1.45 (0.00)&	-2.95 (0.00)&	-0.92 (0.00)	&-0.89 (0.01)	&-0.45 (0.03) \\ \bottomrule[1.5pt]
\end{tabular}}
\end{table}

\begin{table}[h]
\caption{\textbf{Classification results.} The test accuracy of MNNs trained using batch-wise covariance trick on image classification. MNN (R) means ReLU MNN, and MNN (H) represents Heaviside MNN.  
}\label{tab:acc_ll_batch}
\centering
\setlength{\tabcolsep}{.78mm}{
\begin{tabular}{lcccccccccc}
 \toprule[1.5pt]
& \multicolumn{1}{l}{MNN (R)}& \multicolumn{1}{l}{MNN (H)} \\ \midrule
MNIST        & {98.99 (0.00)} &{99.00 (0.00)}     \\
FashionMNIST & {89.66 (0.00)} &{90.13 (0.00)}   \\
CIFAR-10     & {70.75 (0.01)} &{69.86 (0.01)}   \\
CIFAR-100    & {35.84 (0.02)} &{40.21 (0.02)}   \\ \bottomrule[1.5pt]
\end{tabular}}
\end{table}

\section*{\ma{Uncertainty quantification on Heaviside MNNs}}
\begin{figure}[h]
    \centering
    \includegraphics[width=\textwidth]{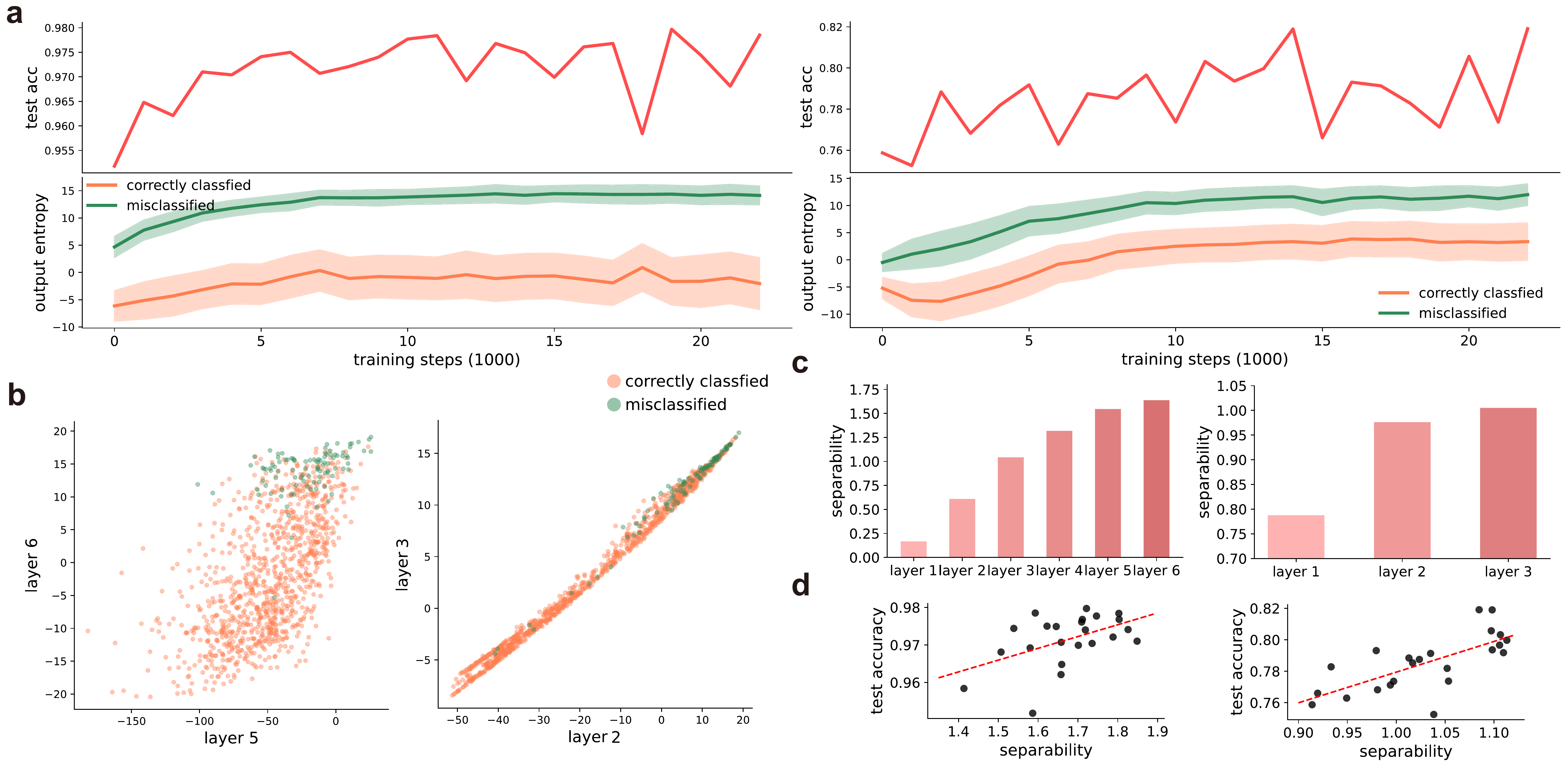}
    \caption{{\bf Emerging covariance of Heaviside MNNs faithfully captures the prediction uncertainty. }(Mixed) Heaviside MNN trained on MNIST (CIFAR-10) for image classification.
    \textbf{a,}
    The training of MNNs on MNIST (left) and CIFAR-10 (right), with the accuracy (red line) and entropy (orange and green for correctly classified inputs and misclassified inputs respectively) on the test set.
    %
    %The shadows indicate the level of half standard deviations. As the training progresses, the entropy of correctly classified inputs and misclassified inputs diverges.
    \textbf{b,}
    The entropy of correctly classified and misclassified inputs in the last two layers of MNIST (left) and CIFAR-10 (right), and the misclassified inputs result in relatively higher entropy in both layers.
    \textbf{c,}
    The entropy separability of correctly classified and misclassified inputs on MNIST (left) and CIFAR-10 (right) increases as the layer index increases.
    \textbf{d,}
    The relationship between the separability and test accuracy during training on MNIST (left) and CIFAR-10 (right) was analyzed using linear regression. %The red dashed line represents the regression line, 
    The coefficients of determination of 0.6803 for MNIST and 0.6068 for CIFAR-10.
    }
    \label{fig:classifier_all_heav}
\end{figure}

We report the uncertainty quantification results of Heaviside MNNs for classification in Fig.~\ref{fig:classifier_all_heav}, which is consistent with that of ReLU MNNs in Fig.~\ref{fig:classifier_all}.

\section*{\ma{More performance comparison with other probabilistic computation methods}}

\begin{table}[h]
\caption{\textbf{Performance on the regression problems of MNN without correlations.} Log-likelihood of different models trained on the UCI regression datasets over random train/test splits. The data shows the mean (standard deviation) over 20 runs.
MNN (R) means ReLU MNN, MNN (H) represents Heaviside MNN. dMMM represents MNN without considering the correlation.
}\label{tab:regression}
\setlength{\tabcolsep}{.78mm}{
\begin{tabular}{lcclccccc}
\toprule[1.5pt]
& boston    & concrete    & \multicolumn{1}{c}{kin8}  & energy  & power  & protein & wine & yacht   \\  \midrule
MNN (R)     & -2.44 (0.00)	&-3.59 (0.01)&	0.92 (0.00)&	-2.30 (0.01)&	-2.92 (0.00)&	-0.97 (0.00)&	-0.75 (0.00)&	-0.29 (0.06)     \\
dMNN (R)    & -2.78 (0.54)	&-3.27 (0.00)&	0.67 (0.00)&	-2.36 (0.03)&	-3.00 (0.00)&	-2.63 (0.25)&	-0.77 (0.00)&	-0.06 (0.30)      \\  \midrule
MNN (H)     &-2.65 (0.03)	&-3.28 (0.01)&	0.71 (0.00)&	-1.31 (0.01)&	-2.94 (0.00)&	-0.92 (0.00)&	-0.90 (0.01)&	-0.40 (0.00)      \\
dMNN (H)    & -2.63 (0.04)	&-3.18 (0.11)&	0.72 (0.00)&	-1.46 (0.01)&	-2.90 (0.00)&	-0.88 (0.00)&	-0.92 (0.01)&	-0.25 (0.04)      \\\bottomrule[1.5pt]
\end{tabular}
}
\end{table}

\begin{table}[h]
\caption{\textbf{Comparison of performance on the classification problems.}  Comparison of test accuracy on image classification. The data shows the mean (standard deviation) over 20
runs.
We compare MNNs (using ReLU or Heaviside activations, with or without correlations) with three representative uncertainty quantification methods. VarOut, Variational dropout~\cite{kingma2015variational}; DVI, deterministic variational inference~\cite{wu2018deterministic}; VBP, Variance Back-Propagation~\cite{haussmann2020sampling}.
}\label{tab:acc_ll}
\centering
\setlength{\tabcolsep}{1.2mm}{
\begin{tabular}{lcccccccccccc}
 \toprule[1.5pt]
&\multicolumn{1}{l}{MNN (R)} &\multicolumn{1}{l}{dMNN (R)}  & \multicolumn{1}{l}{MNN (H)} &\multicolumn{1}{l}{dMNN (H)} & \multicolumn{1}{l}{VarOut} & \multicolumn{1}{l}{DVI} & \multicolumn{1}{l}{VBP}       \\ \midrule
MNIST       &{99.04 (0.00)} &{99.03 (0.02)}  & 99.01 (0.00)       &{ 99.05 (0.00)}   &98.88 (0.05)    &99.03 (0.06) &{\bf99.15 (0.06)}\\
FashionMNIST&{ 89.85 (0.03)}    &{90.05 (0.00)}  &{\bf 90.08 (0.00)}  &{89.93 (0.00)}   &89.19 (0.24)     &89.67 (0.07) &{89.90 (0.16)}\\
CIFAR-10    &{\bf71.04 (0.01)}  &{70.57 (0.48)}  &{70.36 (0.01)}   &{69.63 (0.02)}   &66.60 (1.06)     &64.85 (1.13) &68.67 (1.00)\\
CIFAR-100   &35.56 (0.01)       &{31.01 (0.03)}  &{\bf 40.30 (0.00)}  &{40.04 (0.01)}   &37.15 (1.43)     &33.79 (1.14) &{39.03 (1.61)}\\ \bottomrule[1.5pt]
\end{tabular}
}
\end{table}

We show that our MNNs can achieve comparable or even superior results than representative probabilistic computation methods in terms of model performance.
For regression tasks, we train MNNs on UCI regression datasets~\cite{asuncion2007uci}.
Note that, the MSE loss does not involve output covariance.
We consider MNN using different activations (ReLU and Heaviside function). 
To investigate the effect of correlations, we construct dMMN, in which we ignore all the correlations between neurons at each layer.
We use {\em log-likelihood (LL)} on the test set as the metrics.
As shown in Table.~\ref{tab:regression}, we compare the log-likelihood of various methods with our MMN, dMNN and show that both MNN and dMNN outperform other methods in most datasets.
Additionally, MNN is superior to dMNN for both ReLU and Heaviside functions in most cases, implying that correlation is meaningful.
For classification tasks, we train MNNs through SMUC on MNIST, FashionMNIST~\cite{xiao2017fashion}, CIFAR-10 and CIFAR-100~\cite{krizhevsky2009learning} respectively. We use modified LeNet5 as the architecture of MNNs, following the settings in~\cite{haussmann2020sampling}.
The training loss is the cross entropy, calculated by the prediction mean and the ground-truth label.
As shown in Table.~\ref{tab:acc_ll}, MNNs can achieve comparable performance than other probabilistic computation methods.

\section*{\ma{Theoretical interpretation of the relationship between sensitivity and entropy}}

\begin{figure}[h]
    \centering
    \includegraphics[width=0.8\textwidth]{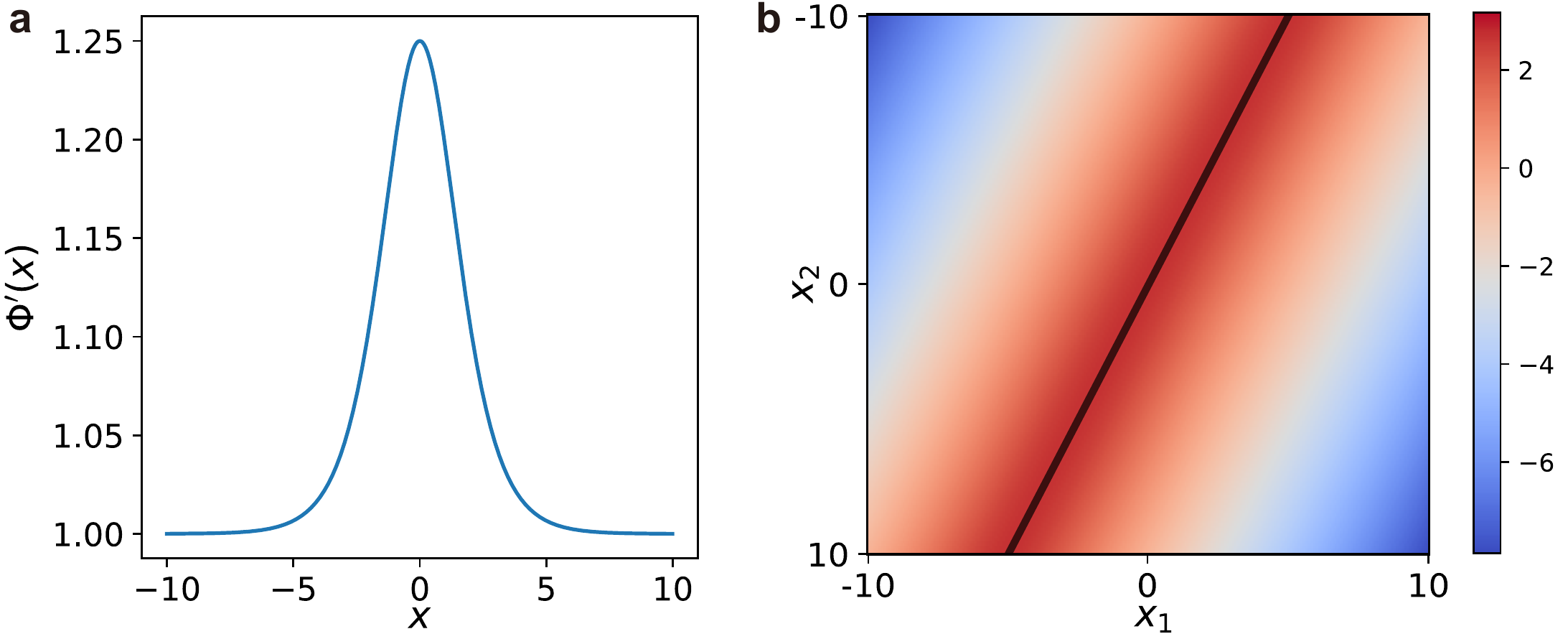}
    \caption{{\bf Entropy of the stochastic network captures the sensitivity}. \textbf{a,} The derivation of the sigmoid function reaches its maximum at $0$. \textbf{b,} We consider a two-dimensional case for illustration. We set $\mathbf{w}=(1,0.5),b=0,C=I$. We calculate the prediction entropy at each $\mathbf{x}\in\mathbb{R}^2$. The closer a point to the decision boundary (black line), the higher the prediction entropy (heatmap).}
    \label{fig:sensi_uncertain_all}
\end{figure}

For illustrative purposes, we analyze a simple binary logistic regression problem to demonstrate how the sensitivity related to the prediction entropy in the stochastic network.
Consider that a binary classification problem, in which $\mathbf{x}\in\mathbb{R}^d$ belongs to the class A if and only if $\mathbf{w}^T\mathbf{x}+b<0$, otherwise it belongs the class B, where $\mathbf{w}\in\mathbb{R}^d$ and $b\in\mathbb{R}$ are ground-truth parameters.
We construct a stochastic network for logistic regression $y = \Phi(\mathbf{w}^T(\mathbf{x}+\mathbf{z})+b) ,\quad \mathbf{z}\sim\mathcal{N}(\mathbf{0},C)$, 
where $\Phi(s)=\frac{1}{1+\exp(-s)}$
is the sigmoid function, $C\in\mathbb{R}^{d\times d}$ is the covariance matrix of the noise vector $\mathbf{z}$.
Although there is no analytical form of the prediction distribution, we can estimate it using a linear approximation
\begin{align}
    y = \Phi(\mathbf{w}^T(\mathbf{x}+\mathbf{z})+b)
    \approx \Phi(\mathbf{w}^T\mathbf{x}+b)+ {\Phi}'(\mathbf{w}^T\mathbf{x}+b) \mathbf{w}^T\mathbf{z},
\end{align}
where ${\Phi}'(s)$ is the derivative of ${\Phi}(s)$.
Hence, we have
\begin{align}
    p(y|\mathbf{x}) \approx \mathcal{N}(\Phi(\mathbf{w}^T\mathbf{x}+b), ({\Phi}'(\mathbf{w}^T\mathbf{x}+b))^2 \mathbf{w}^TC\mathbf{w}),
\end{align}
and its entropy is calculated as
\begin{align}
  H(p(y|\mathbf{x})) \approx \frac{1}{2}(1+\log 2\pi+  \log [ ({\Phi}'(\mathbf{w}^T\mathbf{x}+b))^2 \mathbf{w}^TC\mathbf{w}]),
\end{align}
which is dependent on $\mathbf{x}$.
Noticed that $d(\mathbf{x})=\frac{\mathbf{w}^T\mathbf{x}+b}{\left\|\mathbf{w}\right\|}$ is the (signed) distance from $\mathbf{x}$ to the decision boundary $\mathbf{w}^T\mathbf{x}+b=0$. The approximated entropy can be rewritten as
\begin{align}
  H(p(y|\mathbf{x})) \approx \frac{1}{2}(1+\log 2\pi+  \log[ {\Phi}'(\left\|\mathbf{w}\right\|d(\mathbf{x}))^2 \mathbf{w}^TC\mathbf{w}]). 
\end{align}
Additionally, notice that ${\Phi}'(s) = \frac{1}{1+\exp(-s)} - \frac{1}{(1+\exp(-s))^2}$
reaches its maximum at $s=0$, and monotonically decreases as $s$ goes far away from $0$ (Fig.~\ref{fig:sensi_uncertain_all}\textbf{a}).
Therefore, when $\mathbf{x}$ closer to the decision boundary (Fig.~\ref{fig:sensi_uncertain_all}\textbf{b}), the entropy is higher.
As a result, the entropy of the stochastic model reflects the sensitivity.

\clearpage

%\nolinenumbers
\bibliographystyle{unsrt}  
\bibliography{references} 

\clearpage

\begin{flushleft}
{\Large \textbf{Supplementary Information}\\
\mytitle
}
\newline
{
  Hengyuan Ma\textsuperscript{1},
  Yang Qi\textsuperscript{1,2,3},
  Li Zhang\textsuperscript{4},
  Jie Zhang\textsuperscript{1,2},
  Wenlian Lu\textsuperscript{1,2,5,6,7,8},
  Jianfeng Feng\textsuperscript{1,2,$\ast$} 
\\

\bigskip
\it{1} Institute of Science and Technology for Brain-inspired Intelligence, Fudan University, Shanghai 200433, China
\\
\it{2} Key Laboratory of Computational Neuroscience and Brain-Inspired Intelligence (Fudan University), Ministry of Education, China
\\
\it{3} MOE Frontiers Center for Brain Science, Fudan University, Shanghai 200433, China
\\
\it{4} School of Data Science, Fudan University, Shanghai 200433, China
\\
\it{5} School of Mathematical Sciences, Fudan University, No. 220 Handan Road, Shanghai, 200433, Shanghai, China
\\
\it{6} Shanghai Center for Mathematical Sciences, No. 220 Handan Road, Shanghai, 200433, Shanghai, China
\\
\it{7} Shanghai Key Laboratory for Contemporary Applied Mathematics, No. 220 Handan Road, Shanghai, 200433, Shanghai, China
\\
\it{8} Key Laboratory of Mathematics for Nonlinear Science, No. 220 Handan Road, Shanghai, 200433, Shanghai, China 
\newline
$\ast$ jffeng@fudan.edu.cn 
}
\end{flushleft}

\setcounter{figure}{0}
\renewcommand{\thefigure}{S\arabic{figure}}
\renewcommand{\thetable}{S\arabic{table}}
\setcounter{section}{0}
\setcounter{equation}{0}
\renewcommand{\theequation}{S\arabic{equation}}
\setcounter{page}{1}

\section{Derivation of the moment activations (MAs)}
In this section, we derive moment activations (MAs) from a general stochastic network (Eq.~\eqref{eq:dynamics_original}). First, we employ a mean field approximation to simplify the network dynamics and derive expressions that describe the relationship between the mean and variance of the output signals based on the input signals of the network.
Subsequently, we utilize the linear response approach to estimate the relationship between the covariance of the output signals and the input signals of the network.
Then, we return to the specific stochastic network used in the main paper (Eq.~\eqref{eq:stoc_network}) and provide specific expressions for the corresponding MAs.

\subsection{Mean field approximation for a nonlinear stochastic network}
Consider a nonlinear stochastic neural network
\begin{align}
    &\frac{d\mathbf{x}}{dt} = \mathbf{f}(\mathbf{x}) + W\mathbf{v}+\mathbf{v}_{s},\quad \mathbf{v} = \mathbf{h}(\mathbf{x}) \label{eq:dynamics_original}
\end{align}
where $\mathbf{x}\in\mathbb{R}^N$ is the state of the neurons, $\mathbf{f}(\cdot)=(f_i(\cdot))_{i=1}^N$ is an element-wise function which governs the dynamics of $\mathbf{x}$, $\mathbf{v}$ is the signals produced by the neurons, which are transmitted to other neurons through the weight $W=(w_{ij})_{i,j=1}^N$, $\mathbf{h}(\cdot)=(h_i(\cdot))_{i=1}^N$ is  an element-wise signal function that maps the state of the neurons $\mathbf{x}$ to its output signal $\mathbf{v}$, $\mathbf{v}_{s}$ is the stochasticity of the neurons, which is a Brownian motion with drift $\bm{\mu}_{s}\in\mathbb{R}^N$ and diffusion coefficient $\Lambda\in\mathbb{R}^{N\times N}$.

We assume that the system reaches a stationary distribution, i.e., the distribution of $x_i$ and $v_i$ are independent of time $t$.
Although  this stationary distribution can not be solved analytically in general, we can approximate it via a mean field approximation.
Concretely, we approximate the weighted received signals
$W\mathbf{v}+\mathbf{v}_s$ by a Brownian motion $\bm{\xi}$.
We further assume that the signals $\mathbf{v}$ are independent of the neural stochasticity $\mathbf{v}_{s}$.
Denote the mean and covariance of $\mathbf{v}$ as
\begin{align}
    \bm{\mu} = \mathbb{E} [\mathbf{v}], \quad C = \mathbb{E}[(\mathbf{v} - \bm{\mu})(\mathbf{v} - \bm{\mu})^{\mathrm{T}}].
\end{align}
Then the drift $\bar{\bm{\mu}}$, the diffusion coefficient $\bar{C}$ of $\bm{\xi}$ as
\begin{align}\label{eq:weight_sum}
    \bar{\bm{\mu}} = W\bm{\mu}+\bm{\mu}_{s},\quad
    \bar{C} = WCW^{\mathrm{T}}+\Lambda.
\end{align}
Under this mean field approximation, Eq.~\eqref{eq:dynamics_original} can be rewritten as
\begin{align}\label{eq:diffusion_state}
  \frac{d\mathbf{x}}{dt} = \mathbf{f}(\mathbf{x}) + \bm{\xi}.
\end{align}
Denote the stationary distribution of each $x_i$ as $p(x_i|\bar{\mu}_i,\bar{\sigma}^2_i)$ with $\bar{\sigma}^2_i=\bar{C}_{ii}$ and that of each pair $(x_i,x_j)$ as $p(x_i,x_j|\bar{\mu}_i,\bar{\mu}_j,\bar{\sigma}^2_i,\bar{\sigma}^2_j,\rho_{ij})$, where $\rho_{ij} = \bar{C}_{ij}/(\bar{\sigma}_i\bar{\sigma}_j)$ is the correlation coefficient of the input signals of two neurons.
The distribution $p(x_i|\bar{\mu}_i,\bar{\sigma}^2_i)$ is the solution to the time-independent Fokker-Planck equation~\cite{gardiner1985handbook}
% %
\begin{align}\label{eq:fokker1}
    \frac{\partial}{\partial x_i}\big((f_i(x_i)+\bar{\mu}_i)p(x_i)\big)=\bar{\sigma}^2_i\frac{\partial^2}{\partial x_i^2}p(x_i).
\end{align}
and the joint distribution $p(x_i,x_j|\bar{\mu}_i,\bar{\mu}_j,\bar{\sigma}^2_i,\bar{\sigma}^2_j,\bar{C}_{ij})$ is the solution to the time-independent Fokker-Planck 
\begin{align}\label{eq:fokker2}
(\frac{\partial}{\partial x_i}(f_i(x_i)+\bar{\mu}_i)+\frac{\partial}{\partial x_j}(f_i(x_j)+\bar{\mu}_j))p(x_i,x_i)=(\bar{\sigma}^2_i\frac{\partial^2}{\partial x_i^2}+\bar{\sigma}^2_j\frac{\partial^2}{\partial x_i^2}+\bar{C}_{ij}\frac{\partial^2}{\partial x_i\partial x_j})p(x_i,x_j).
\end{align}
After solving Eq.~\eqref{eq:fokker1}-\eqref{eq:fokker2}, we can calculate the first two cumulants of the output signals $\mathbf{v}$
\begin{align}\label{eq:original_map1}
    &\mu_i =\mathbb{E} [v_i] =  \int h_i(x_i)p(x_i|\bar{\mu}_i,\bar{C}_i)dx_i,\\
    &
    C_{ij} =
    \mathrm{cov}[v_i,v_j]
    = \int (h_i(x_i)-\mathbb{E} [v_i])(h_j(x_j)-\mathbb{E} [v_j])p(x_i,x_j|\bar{\mu}_i,\bar{\mu}_j,\bar{\sigma}^2_i,\bar{\sigma}^2_j,\bar{C}_{ij})dx_idx_j.\label{eq:original_map2}
\end{align}
We call Eq.~\eqref{eq:original_map1}-\eqref{eq:original_map2} as the {\em mean activation} and {\em covariance activation} respectively, and they constitute the {\em moment activations (MAs)}. 

The computation complexity of the double integration in the covariance activation (Eq.\eqref{eq:original_map2}) is high. To address this, we simplify the covariance activation using the linear response theory~\cite{de2007correlation}.
The covariance between the output signals of two neurons 
$\mathrm{cov}[v_i,v_j]$ can be considered as a function of $\rho_{ij}$, denoted as $ C_{v,ij}(\rho_{ij})$,
Assuming that $\rho_{ij}$ is small, we can approximate $C_{v,ij}(\rho_{ij})$ using a first-order Taylor expansion
\begin{align}\label{eq:original_map3}
    C_{v,ij}(\rho_{ij})\sim \frac{d}{d\rho_{ij}}C_{v,ij}(0)\rho_{ij},
\end{align}
where the derivative $\frac{d}{d\rho_{ij}}C_{v,ij}(0)$ can be considered as a sort of susceptibility describing how two neurons transfer weak correlated input signals~\cite{de2007correlation}. 

\subsection{Moment activations for Ornstein–Uhlenbeck processes}
We now develop the MAs for the Ornstein–Uhlenbeck system (Eq.~\eqref{eq:stoc_network}) in the main paper, where $\mathbf{f}(\mathbf{x})=-\mathbf{x}$.
In this case, the stationary distribution $p(x_i|\bar{\mu}_i,\bar{\sigma}^2_i)$ is a Gaussian distribution 
with density
\begin{align}
     \frac{1}{\sqrt{2\pi }\bar{\sigma}_i}\exp\{-\frac{(x_i-\bar{\mu}_i)^2}{2\bar{\sigma}_i^2}\}
\end{align}
and $p(x_i,x_j|\bar{\mu}_i,\bar{\mu}_i,\bar{\sigma}^2_i,\bar{\sigma}^2_j,\bar{C}_{ij})$ is a two-dimensional Gaussian distribution with density
\begin{align}
\frac{1}{2\pi \sqrt{1-\rho_{ij}^2}\bar{\sigma}_i\bar{\sigma}_j} \exp(-\frac{1}{2(1-\rho_{ij}^2)\bar{\sigma}_i^2\bar{\sigma}_j^2}(x_i-\bar{\mu}_i ,x_j-\bar{\mu}_j)\begin{pmatrix}
\bar{\sigma}_j^2&-\rho_{ij}\bar{\sigma}_i\bar{\sigma}_j\\
-\rho_{ij}\bar{\sigma}_i\bar{\sigma}_j&\bar{\sigma}_i^2
\end{pmatrix}(x_i-\bar{\mu}_i ,x_j-\bar{\mu}_j)^\mathrm{T}).
\end{align}
For convenience, we will denote $\bar{\sigma}^2_i$ as $\bar{C}_i$
Using Eq.~\eqref{eq:original_map1}-\eqref{eq:original_map2}, we have the mean and variance of the output signals $v_i$
\begin{align}
    &m_{v,i}(\bar{\mu}_i,\bar{C}_i) := \mathbb{E}[v_i] =  \int h_i(x_i)p(x_i|\bar{\mu}_i,\bar{C}_i)dx_i,\label{eq:mean_si}\\
    &C_{v,i}(\bar{\mu}_i,\bar{C}_i) := \mathbb{E}[(v_i -\mathbb{E}[v_i] )^2] =  \int h_i^2(x_i)p(x_i|\bar{\mu}_i,\bar{C}_i)dx_i - m_v(\bar{\mu}_i,\bar{C}_i)^2,\label{eq:var_si}
\end{align}
Using Eq.~\eqref{eq:original_map3},
the approximated covariance is
\begin{align}
   C_{v,ij}(\bar{\mu}_i,\bar{\mu}_j,\bar{C}_i,\bar{C}_j,\bar{C}_{ij})\approx \frac{1}{2\pi}\int h_i(\sigma_ix_i+\mu_i) \exp(-\frac{1}{2}x_i^2)x_idx_i\int  h_j(\sigma_jx_j+\mu_j)\exp(-\frac{1}{2}x_j^2)x_jdx_j\rho_{ij},
\end{align}
which can be rewritten as
\begin{align}
    C_{v,ij}(\bar{\mu}_i,\bar{C}_i,\bar{\mu}_j,\bar{C}_j,c) =  \chi_i(\bar{\mu}_i,\bar{C}_i)\chi_j(\bar{\mu}_j,\bar{C}_j)\rho_{ij},\label{eq: cov_map}
\end{align}
where
\begin{align}\label{eq:cov_si}
      \chi_i(\bar{\mu}_i,\bar{C}_i) = \frac{1}{\sqrt{2\pi} }\int h_i(\sqrt{\bar{C}_i}x+\bar{\mu}_i) e^{-\frac{x^2}{2}}xdx.
\end{align}

\section{Theoretical properties of supervised mean unsupervised covariance (SMUC)}\label{sec:heaviside_mean_variance}
We train an MNN $\tilde{M}$ through SMUC for learning the output mean and covariance of the ground-truth MNN $M_{GT}$. 
We prove the following three theoretical properties about the SMUC under mild conditions:
\begin{itemize}
    \item \ma{SMUC is equivalent to the stochastic Riemannian gradient descent, and the parameter of the $\tilde{M}$ converges to the point where the   loss gradient is vanished(Thm.~\ref{thm:rgd})},
    \item $\tilde{M}$ can learn ground-truth mean and variance at each layer even if the ground-truth hyper-parameters $\sigma_l$ are unknown \ma{(Thm.~\ref{thm:mean_var}),}
    \item As the number of the layers goes to infinity, the output covariance of $\tilde{M}$ converges to that of $M_{GT}$ \ma{(Thm.~\ref{thm:cov}).}
\end{itemize}
We begin by analyzing the case where the network has only one nonlinear layer with Heaviside activations in Sec.~\ref{sec:srgd}-\ref{sec:variance}, and then generalize the results to the multiple layer cases in Sec.~\ref{sec:multiple_layer}. Next, we analyze the covariance of MNN in Sec.~\ref{sec:cov_appro}. Finally, we generalize the above results to case with general nonlinear activation functions in Sec.~\ref{sec:general_act}.

Consider the network with one nonlinear layer as follows
\begin{align}\label{eq: oracle_proc}
\begin{aligned}
    \frac{d\mathbf{x}^{(1)}}{dt} &= -\mathbf{x}^{(1)} +\mathbf{x} +\sqrt{2}C_{\mathrm{in}}^{\tfrac{1}{2}}\bm{\xi}^{(1)},\quad \mathbf{v}^{(1)}=\mathbf{x}^{(1)},\\
    \frac{d\mathbf{x}^{(2)}}{dt} &= -\mathbf{x}^{(2)} + W\mathbf{v}^{(1)}+\mathbf{b}+\sqrt{2}\sigma\bm{\xi}^{(2)},\quad \mathbf{v}^{(2)} = \bm{\Theta}(\mathbf{x}^{(2)})\\
    \frac{d\mathbf{y}}{dt}&= -\mathbf{y} + \mathbf{v}^{(2)}
    \end{aligned}
\end{align}
where $\mathbf{x}=(x_i)_{i=1}^m\in \mathbb{R}^m$ is the input, $\mathbf{y}=(y_k)_{k=1}^n\in\mathbb{R}^n$ is the output, 
$\bm{\Theta}(\cdot)$ is the element-wise Heaviside function, $W=(\mathbf{w}_k^{\mathrm{T}})_{k=1}^n$ with $\mathbf{w}_k^{\mathrm{T}}\in\mathbb{R}^m$ are the ground-truth weights, $\mathbf{b}=(b_k)_{k=1}^n\in\mathbb{R}^n$ are the ground-truth biases, $\bm{\xi}^{(l)}$ are unit Gaussian white noises, $\sigma^2$ is the constant ground-truth diffusion coefficient, and $C_{\mathrm{in}}\in \mathbb{R}^{m\times m}$ is the unknown diffusion matrix (positive definite). 
\ma{We denote the corresponding MNN of this stochastic neural network as $M_{GT}$.}
Note that here the covariance matrix of the noise added to the input $\mathbf{x}$ can be arbitrary positive definite matrix, no need to be supposed as a scalar matrix as that in Eq.~\eqref{eq:stoc_network} in the main paper. In this way, the result derived here can be directly generalized to the cases of multiple nonlinear layer later.
Under the mean field approximation, we simplify the above stochastic via mean field approximation and obtain
\begin{align}\label{eq: oracle}
    y_k=\Theta(\mathbf{w}_k^{\mathrm{T}}[\mathbf{x}+C_{\mathrm{in}}^{\tfrac{1}{2}}\mathbf{z}]+b_k+\sigma u_k),\quad \mathbf{z}\sim\mathcal{N}(\mathbf{0},I_m),\quad u_k\sim\mathcal{N}(0,1) ,\quad k=1,\ldots,n.
\end{align}
For convenience, we introduce an augmented input $\mathbf{x}'=(\mathbf{x},1)\in\mathbb{R}^{m+1}$, augmented weights $\mathbf{w}'_k=(\mathbf{w}_k,b_k)\in\mathbb{R}^{m+1}$ which include the bias terms as their component, augmented covariance matrix
\begin{align}
    \begin{pmatrix}
 C_{\mathrm{in}} & \mathbf{0}  \\
  \mathbf{0}^{\mathrm{T}}& 0
\end{pmatrix},
\end{align}
and augmented noise term $\mathbf{z}'=(\mathbf{z},z)$ with $z\sim\mathcal{N}(0,1)$ and write
\begin{align}\label{eq: label_y}
    y_k=\Theta(\mathbf{w}_k^{\mathrm{T}}[\mathbf{x}+C_{\mathrm{in}}^{\tfrac{1}{2}}\mathbf{z}]+\sigma u_k),\quad \mathbf{z}\sim\mathcal{N}(\mathbf{0},I_m),\quad u_k\sim\mathcal{N}(0,1),\quad k=1,\ldots,n,
\end{align}
where we use the original notations and do not increase the dimension $m$ by $1$ for convenient.
Using the definition of the moment mapping and Eq.~\eqref{eq: label_y}, we have the ground-truth mean and covariance (variance) are
\begin{align}
\begin{aligned}\label{eq:truth_network}
     &\mu_{y,k}(\mathbf{x}) = m_v(\bar{\mu}_k,\bar{C}_k), \quad C_{y,k}(\mathbf{x}) = C_v(\bar{\mu}_k,\bar{C}_k),\quad k=1,\ldots,n,\\
    &C_{y,kj}(\mathbf{x}) =  \chi(\bar{\mu}_k,\bar{C}_k)\chi(\bar{\mu}_j,\bar{C}_j)c_{kj}, \quad k,j = 1,\ldots,n,\quad k\neq j,
\end{aligned}
\end{align}
with
\begin{align}
    \bar{\mu}_k = \mathbf{w}_k^{\mathrm{T}}\mathbf{x}, \quad 
    \bar{C}_{kj} = \mathbf{w}_k^{\mathrm{T}}C_{\mathrm{in}}\mathbf{w}_j+\delta_{kj}\sigma^2,
    \quad
    \bar{C}_k = \bar{C}_{kk},
    \quad 
    c_{kj} = \frac{\bar{C}_{kj}}{\sqrt{\bar{C}_k\bar{C}_j}},\label{eq:real_C}
\end{align}
where $\delta_{kj}=1$ if $k=j$ else $\delta_{kj}=0$.
To learn the mapping $\mathbf{x}\mapsto q(\mathbf{y}|\mathbf{x})$ (Eq.~\eqref{eq: oracle}) up to its second cumulants, we construct a single layer MNN denoted as $\tilde{M}$ in which the feed-forward propagation for an input $\mathbf{x}$ is calculated as
\begin{align}
\begin{aligned}
    &\tilde{m}_{y,k}(\mathbf{x}) = m_v(\tilde{\mu}_k,\tilde{C}_k), \quad \tilde{C}_{y,k}(\mathbf{x}) = C_v(\tilde{\mu}_k,\tilde{C}_k),\quad k=1,\ldots,n,\\
    &\tilde{C}_{y,kj}(\mathbf{x}) =  \chi(\tilde{\mu}_k,\tilde{C}_k)\chi(\tilde{\mu}_j,\tilde{C}_j)\tilde{c}_{kj}, \quad k,j = 1,\ldots,n,\quad k\neq j,
\end{aligned}
\end{align}
with
\begin{align}
    \tilde{\mu}_k = \tilde{\mathbf{w}}_k^{\mathrm{T}}\mathbf{x}, \quad 
    \tilde{C}_{kj} = \tilde{\mathbf{w}}_k^{\mathrm{T}}\tilde{C}_{\mathrm{in}}\tilde{\mathbf{w}}_j+\delta_{kj}\tilde{\sigma}^2,
    \quad
    \tilde{C}_k = \tilde{C}_{kk},
    \quad 
    \tilde{c}_{kj} = \frac{\tilde{C}_{kj}}{\sqrt{\tilde{C}_k\tilde{C}_j}}\label{eq:model_summation}.
\end{align}
where $\tilde{y}$ is the predicted $y$ using its mean and $(\tilde{\mathbf{w}}_1,\ldots,\tilde{\mathbf{w}}_n)$ are the weights to be learned, $\tilde{C}_{\mathrm{in}}$ and $\tilde{\sigma}$ are the manually set, which may be different from $C_{\mathrm{in}}$ and $\sigma$.
Note that in the case of Eq.~\eqref{eq:mnn_feedforward} of the main paper, we set $\tilde{C}_{\mathrm{in}}$ as $\sigma_1^2I$ and $\sigma$ here corresponds to $\sigma_2$.
The choice of $\tilde{C}_{\mathrm{in}}$ is arbitrary, as long as it is positive definite.
Denote the objective function as loss $ \mathcal{L}(\mathbf{y},\bm{\tilde{\mu}}_y)$, where $\bm{\tilde{\mu}}_y = (\tilde{m}_{y,k})_{k=1}^m$. 
The training procedure of SMUC is 
\begin{align}\label{eq: update_w}
    \tilde{\mathbf{w}}_{k,t+1} = \tilde{\mathbf{w}}_{k,t}- \gamma_{t}  \frac{\partial \mathcal{L}}{\partial  \tilde{\mathbf{w}}_{k,t}},\quad,k=1,\dots,n,\quad t=1,\ldots,T,
\end{align}
where $\gamma_t$ is the learning rate, and $T$ is the step number of training.
Once again, we would like to emphasize that all second cumulants are treated as constants during gradient calculation and do not participate in backpropagation.

Using the MAs, we can approximate $\mathbf{y}$ by a Gaussian distribution using its first two cumulants as $ \mathbf{y} \sim \bm{\mu}_{y}+ C_y^{\tfrac{1}{2}} \mathbf{z},\quad \mathbf{z}\sim \mathcal{N}(\mathbf{0},I)$,
where $\bm{\mu}_{y} = (\mu_{y,k})_{k=1}^n$, and $C_{y} = (C_{y,kj})_{k,j=1}^n$.
Using this approximation, the objective can be rewritten as
\begin{align}\label{eq:loss_approximate}
    \mathcal{L}(\mathbf{y},\bm{\tilde{\mu}}_y) \sim  \mathcal{L}(\bm{\mu}_{y},\bm{\tilde{\mu}}_y) + J(\bm{\mu}_{y},\bm{\tilde{\mu}}_y)C_y^{\tfrac{1}{2}}\mathbf{z},
\end{align}
where $J(\bm{\mu}_{y},\bm{\tilde{\mu}}_y)$ is the Jacobian matrix of the function $\mathcal{L}(\mathbf{y},\bm{\tilde{\mu}}_y)$ at $\mathbf{y}=\bm{\mu}_{y}$.
Then the gradient at the $t$ step can be approximately rewritten as the summation of the deterministic part $\frac{\partial}{\partial \tilde{\mathbf{w}}_{k,t} }\mathcal{L}(\bm{\mu}_{y},\bm{\tilde{\mu}}_y)$
with a stochastic term $    \bm{\xi}_{k,t}=\frac{\partial \mathcal{L}}{\partial  \tilde{\mathbf{w}}_{k,t}} J(\bm{\mu}_{y},\bm{\tilde{\mu}}_y)C_y^{\tfrac{1}{2}}\mathbf{z}$.

\subsection{\ma{SMUC is equivalent to stochastic Riemannian gradient descent}}\label{sec:srgd}

\ma{We now prove the Thm.~\ref{thm:rgd} in the single-layer case.
}
\begin{proof}
We first prove the equivalence between SMUC and the stochastic Riemannian gradient descent. The main thread of our proof is as follows. We leverage the scalar property of the Heaviside MAs (Eq.~\eqref{eq:scalar_property}) to reformulate the update equation for the SMUC (Eq.~\eqref{eq: update_w}).
This reformulation enables us to align the variances of each layer in the MNN with the variances in the ground-truth network (Eq.~\eqref{eq:truth_network}) during the gradient descent process on the mean.
Furthermore, the alignment operation can be viewed as projecting the parameters onto a manifold where the variance matches the ground-truth value. By employing this alignment technique, we establish a correspondence between the SMUC and stochastic Riemannian gradient descent.

Recall that according to Eq.~\eqref{eq:heav_mapping}, the mean activation of Heaviside function satisfies the scalar property
\begin{align}\label{eq:scalar_property}
    m_v(\mu,C)  = m_v(a\mu,a^2C) ,\quad \forall a> 0,\mu\in\mathbb{R},C>0.
\end{align} 
This suggests that the mean mapping remains unchanged when simultaneously scaling the mean by a factor of $a$ and the covariance by a factor of $a^2$.  
Denoting a scaling factor $\alpha_{k,t} =\sqrt{\frac{\bar{C}_k}{\tilde{C}_{k,t}}}$, where $\tilde{C}_{k,t}$
is defined in Eq.~\eqref{eq:model_summation},
we can rewrite the loss $\mathcal{L}$ as 
\begin{align}
\mathcal{L}(\bm{\mu}_{y},\bm{\tilde{\mu}}_y)= \mathcal{L}\big((m_v(\bar{\mu}_k,\bar{C}_k))_{k=1}^n,(m_v(\tilde{\mathbf{w}}_{k,t}^{\mathrm{T}}\mathbf{x}, \tilde{C}_k))_{k=1}^n \big)=\mathcal{L}\big((m_v(\bar{\mu}_k,\bar{C}_k))_{k=1}^n,(m_v(\hat{\mathbf{w}}_{k,t}^{\mathrm{T}}\mathbf{x}, \bar{C}_k))_{k=1}^n \big),\label{eq:iso_variance_eq}
\end{align}
where we define $\hat{\mathbf{w}}_{k,t}=\alpha_{k,t} \tilde{\mathbf{w}}_{k,t}$.
Hence, the gradient can be rewritten as
\begin{align}\label{eq:substitude_w}
\frac{\partial}{\partial \tilde{\mathbf{w}}_{k,t}}\mathcal{L}(\bm{\mu}_{y},\bm{\tilde{\mu}}_y)
    =
\alpha_{k,t}\frac{\partial}{\partial \hat{\mathbf{w}}_{k,t}}\mathcal{L}\big((m_v(\bar{\mu}_k,\bar{C}_k))_{k=1}^n,(m_v(\hat{\mathbf{w}}_{k,t}^{\mathrm{T}}\mathbf{x}, \bar{C}_k))_{k=1}^n \big).
\end{align}
As a result, we can write the update equation Eq.~\eqref{eq: update_w} as
\begin{align}
    \alpha_{k,t}\tilde{\mathbf{w}}_{k,t+1} &= \hat{\mathbf{w}}_{k,t}- \gamma_{t} [\alpha^2_{k,t}\frac{\partial}{\partial \hat{\mathbf{w}}_{k,t}}\mathcal{L}\big((m_v(\bar{\mu}_k,\bar{C}_k))_{k=1}^n,(m_v(\hat{\mathbf{w}}_{k,t}^{\mathrm{T}}\mathbf{x}, \bar{C}_k))_{k=1}^n \big) + \alpha_{k,t}\bm{\xi}_{k,t}]
\end{align}
If we track the weights $\hat{\mathbf{w}}_{k,t}$ instead of $\tilde{\mathbf{w}}_{k,t}$ at each step $t$, the update of weight in Eq.~\eqref{eq: update_w} is equivalent to two steps
\begin{align}
    &\breve{\mathbf{w}}_{k,t+1} = \hat{\mathbf{w}}_{k,t}- \gamma_t (\alpha^2_{k,t} \frac{\partial}{\partial \hat{\mathbf{w}}_{k,t} } \mathcal{L}_{\mathrm{mean}}
    + \alpha_{k,t}\bm{\xi}_{k,t})\label{eq:start_update_alpah}\\
    &\hat{\mathbf{w}}_{k,t+1} =\frac{\alpha_{k,t+1}}{\alpha_{k,t} }\breve{\mathbf{w}}_{k,t+1},\label{eq:end_update_alpah}
\end{align}
where $\mathcal{L}_{\mathrm{mean}}$ is the loss on the mean taking the second cumulants as that of $M_{GT}$
\begin{align}
        \mathcal{L}_{\mathrm{mean}} =\mathcal{L}\big((m_v(\bar{\mu}_k,\bar{C}_k))_{k=1}^n,(m_v(\hat{\mathbf{w}}_{k,t}^{\mathrm{T}}\mathbf{x}, \bar{C}_k))_{k=1}^n \big).
\end{align}

Next, we show that Eq.~\eqref{eq:start_update_alpah} and Eq.~\eqref{eq:end_update_alpah} is equivalent to the \textit{stochastic Riemannian gradient descent}~\cite{boumal2023introduction}.
Define the manifold $\mathcal{M}=\mathcal{M}_1\otimes\mathcal{M}_2\otimes\cdots\otimes\mathcal{M}_n\in \mathbb{R}^{m\times n}$, where $\otimes$ is the Cartesian product, and
\begin{align}\label{eq:iso_uncer}
   \mathcal{M}_{k}=\{\hat{\mathbf{w}}\in\mathbb{R}^m, \hat{\mathbf{w}}^\mathrm{T}\tilde{C}_{\mathrm{in}}\hat{\mathbf{w}}+\alpha^2_{k,t}\tilde{\sigma}^2 =\bar{C}_k\},\quad k=1,\ldots,n.
\end{align}
Due to the definition of $\alpha^2_{k,t}$, we always have $\alpha^2_{k,t}\tilde{\sigma}^2 <\bar{C}_k$,
hence $\mathcal{M}_{k}$ is not empty.
We call $\mathcal{M}$ as iso-variance sphere, as the weights $(\hat{\mathbf{w}}_k)$ on it ensure that the mean activation $m_v(\cdot,\cdot)$ receives the ground-truth variance $\bar{C}_k$, as described in Eq.~\eqref{eq:real_C}.
At each iteration, we first update the parameters to minimize $\mathcal{L}_{\mathrm{mean}}$ through stochastic gradient Langevin dynamics~\cite{welling2011bayesian} in the Euclidean space,
and then we project the parameters back to the manifold $\mathcal{M}$ (Eq.~\eqref{eq:end_update_alpah}). 
Thus, Eq.~\eqref{eq:start_update_alpah} and Eq.~\eqref{eq:end_update_alpah} is equivalent to a stochastic Riemannian gradient descent on the iso-variance sphere $\mathcal{M}$
\begin{align}\label{eq:srgd}
    \hat{\omega}_{t+1} = R_{\hat{\omega}_{t}}[- \gamma_t( \alpha_{t}^2 \odot \frac{\partial}{\partial \hat{\omega}_{t} } \mathcal{L}_{\mathrm{mean}} + \alpha_{t}\odot\bm{\xi}_{t})],
\end{align}
where $\hat{\omega}_{t}\in\mathbb{R}^{m\times n}$ is the concatenation of $(\hat{\mathbf{w}}_{k,t},k=1,\ldots,n)$, $\bm{\xi}_{t}$ is the concatenation of $(\bm{\xi}_{k,t},k=1,\ldots,n)$, $R_{\hat{\omega}}$ is a retraction on $\mathcal{M}$~\cite{boumal2023introduction}, $\alpha_{t}$ is the concatenation of $(\alpha_{k,t},k=1,\ldots,n)$, and $\odot$ is the block-wise multiplication defined as 
\begin{align}
    \alpha_{t}^2 \odot \frac{\partial }{\partial \hat{\omega}_{t}}\mathcal{L}_{\mathrm{mean}} = (\alpha^2_{k,t} \frac{\partial}{\partial \hat{\mathbf{w}}_{k,t} } \mathcal{L}_{\mathrm{mean}},k=1,\ldots,n),\quad  \alpha_{t} \odot \bm{\xi}_{t} =(\alpha_{k,t}\bm{\xi}_{k,t} ,k=1,\ldots,n).
\end{align}
The retraction $R_{\hat{\omega}}$ is a first-order approximation of the corresponding exponential map $\exp_{\hat{\omega}}$ on $\mathcal{M}$~\cite{boumal2023introduction}, and the update Eq.~\eqref{eq:srgd} is a first-order approximation of the following process
\begin{align}\label{eq:srgd2}
    \hat{\omega}_{t+1} = \exp_{\hat{\omega}_{t}}[- \gamma_t( \alpha^2_{k}\odot \frac{\partial}{\partial \hat{\omega}_{t} } \mathcal{L}_{\mathrm{mean}} + \alpha_{t}\odot\bm{\xi}_{t})].
\end{align}

\ma{Next, we prove the convergence of SMUC.} 
We follow the methodology outlined in~\cite{bonnabel2013stochastic} to investigate the convergence of Eq.~\eqref{eq:srgd2}.
Denote $H(\hat{\omega}_{t},\bm{\xi}_{t}) =  \frac{\partial}{\partial \hat{\omega}_{t} } \mathcal{L}_{\mathrm{mean}} +\bm{\xi}_{t} \oslash  \alpha_{t}$, where $\oslash$ is the block-wise division defined as $   \bm{\xi}_{t} \oslash  \alpha_{t} =(\frac{\bm{\xi}_{k,t}}{\alpha_{k,t}} ,k=1,\ldots,n)$.
\ma{
Supposed that we select the proper learning rate $\{\gamma_t\}$ such that it satisfies the following modified usual condition:
\begin{align}\label{eq:usu_cond}
\sum_{t}\gamma_t^2\alpha_{k,t}^4 < +\infty,\quad\sum_{t} \gamma_t\alpha_{k,t}^2 = +\infty,\quad \forall k=1,\ldots,n .
\end{align}}
Since $\mathcal{M}$ is bounded, there exists a compact set $K\subset\mathbb{R}^m$, such that $\hat{\omega}_{t}\in K,\quad\forall t>0$.
Since $\mathcal{M}$ is the product of spheres $\mathcal{M}_k$ embedded in the Euclidean space, $\mathcal{M}$ is a Hadamard manifold, and there is a point $\nu\in\mathcal{M}$ and $S>0$, such that 
\begin{align}
    \inf_{\omega:D(\omega,\nu)>S}\langle \exp_{\omega}^{-1}(\nu),\frac{\partial}{\partial \omega } \mathcal{L}_{\mathrm{mean}} \rangle < 0,
\end{align}
where $D(\omega,\nu)$ is the squared geodesic
distance of $\omega$ and $\nu$ on $\mathcal{M}$.
Additionally, there exists a lower bound on the sectional curvature denoted by $\kappa<0$.
Since $\mathcal{M}$ is bounded and $\bm{\xi}_{k,t}$ are Gaussian, there exists a continuous function $f:\mathcal{M}\rightarrow \mathbb{R}$ that satisfies
\begin{align*}
\begin{aligned}
    f(\omega)^2 \geq \max\{1,\mathbb{E}_{\xi}[\left\| H(\omega,\xi) \right\|^2(1+\sqrt{|\kappa|}(\sqrt{D(\omega,\nu)}+\left\| H(\omega,\xi) \right\|))],
    \mathbb{E}_{\xi}[(2\left\| H(\omega,\xi) \right\|\sqrt{D(\omega,\nu)}+\left\| H(\omega,\xi) \right\|^2)^2]\}.
\end{aligned}
\end{align*}
Then, according to Thm.3 of~\cite{bonnabel2013stochastic}, the iterations
\begin{align}\label{eq:srgd3}
    \hat{\omega}_{t+1} = \exp_{\hat{\omega}_{t}}[- \frac{\gamma_t}{f(\hat{\omega}_{t})}(\alpha^2_{t}\odot \frac{\partial}{\partial \hat{\omega}_{t} } \mathcal{L}_{\mathrm{mean}} + \alpha_{t}\odot\bm{\xi}_{t})].
\end{align}
converges to the point where the gradient $\frac{\partial}{\partial \hat{\omega}_{t}}\mathcal{L}_{\mathrm{mean}}$ vanishes almost surely.
\end{proof}

We notice that when $\tilde{\mathbf{w}}_{k,t}^\mathrm{T}\tilde{C}_{\mathrm{in}}\tilde{\mathbf{w}}_{k,t}>\bar{C}_k$, the step-size $\alpha_{t,k}\gamma_t$ is reduced; when $\tilde{\mathbf{w}}_{k,t}^\mathrm{T}\tilde{C}_{\mathrm{in}}\tilde{\mathbf{w}}_{k,t}<\bar{C}_k$, the step-size $\alpha_{t,k}\gamma_t$ is amplified.
Hence $\alpha_{k,t}$ can be considered as a regularizer: when $\tilde{M}$ underestimates the variance ($\tilde{\mathbf{w}}_{k,t}^\mathrm{T}\tilde{C}_{\mathrm{in}}\tilde{\mathbf{w}}_{k,t}>\bar{C}_k$), $\alpha_{k,t}$ slows down the gradient descent; when the $\tilde{M}$ overestimate the variance ($\tilde{\mathbf{w}}_{k,t}^\mathrm{T}\tilde{C}_{\mathrm{in}}\tilde{\mathbf{w}}_{k,t}<\bar{C}_k$), $\alpha_{k,t}$ accelerates the gradient descent.

\subsection{MNNs learn ground-truth mean and variance}\label{sec:variance}

\ma{
We now prove the Thm.~\ref{thm:mean_var} in the single-layer case.
}
\begin{proof}
After convergence, the trained weight $\tilde{\mathbf{w}}_k$ of $\tilde{M}$ satisfies
\begin{align}\label{eq:real_mean_eq}
    \tilde{m}_{y,k}(\mathbf{x})=m_{v}(\tilde{\mathbf{w}}_k^\mathrm{T}\mathbf{x},\tilde{\mathbf{w}}_k^T\tilde{C}_{\mathrm{in}}\tilde{\mathbf{w}}_k+\tilde{\sigma}^2)=m_v(\alpha_k\mathbf{w}_k^{\mathrm{T}}\mathbf{x},\mathbf{w}_k^\mathrm{T}C_{\mathrm{in}} \mathbf{w}_k+\sigma^2) =m_{v}(\bar{\mu}_k,\bar{C}_k)=\mu_{y,k},
\end{align}
where $\alpha_k =\sqrt{\frac{\bar{C}_k}{\tilde{C}_k}}$.
Therefore, the $\tilde{M}$ can fit the mean of the ground-truth MNN $M_{GT}$. 
For the variance activation of the Heaviside function (Eq.~\eqref{eq:variance_mapping}), we also have the scalar property
\begin{align}
    C_v(\mu,C)  =  C_v(a\mu,a^2C) ,\quad \forall a> 0,\mu\in\mathbb{R},C>0. 
\end{align}
Therefore, similar as Eq.~\eqref{eq:real_mean_eq}, we have
\begin{align}
     \tilde{C}_{y,k}(\mathbf{x})=C_{v}(\tilde{\mathbf{w}}_k^\mathrm{T}\mathbf{x},\tilde{\mathbf{w}}_k^T\tilde{C}_{\mathrm{in}}\tilde{\mathbf{w}}_k+\tilde{\sigma}^2)=C_v(\alpha_k\mathbf{w}_k^{\mathrm{T}}\mathbf{x},\mathbf{w}_k^\mathrm{T}C_{\mathrm{in}} \mathbf{w}_k+\sigma^2) =C_{v}(\bar{\mu}_k,\bar{C}_k)=C_{y,k},
\end{align}
and the $\tilde{M}$ can also fit the variance of $M_{GT}$. 
\end{proof}

\subsection{Multiple-layer cases}\label{sec:multiple_layer}
We extend the previous results about that $\tilde{M}$ can learn the mean and variance of $M_{GT}$ from the case of a single non-linear layer to the multi-layer case. 
Consider the network described in Eq.~\eqref{eq:stoc_network}, with $\mathbf{h}(\cdot)$ is specific as $\bm{\Theta}(\cdot)$, and $\sigma_{1}$ is replaced by $C_{\mathrm{in}}^{\tfrac{1}{2}}$.
Under the mean field approximation, we absorb all the bias terms and noise terms as we did in Sec.~\ref{sec:heaviside_mean_variance} to obtain the corresponding MNN $M_{GT}$

\begin{align}\label{eq: label_y2}
\begin{aligned}
\mathbf{v}^{(1)} &=  \mathbf{x}+C_{\mathrm{in}}^{\frac{1}{2}}\mathbf{z}_{\mathrm{in}}\\
\mathbf{v}^{(l)}&=\bm{\Theta}(W^{(l-1)}\mathbf{v}^{(l-1)}+\sigma_{l}\mathbf{z}^{(l)}),\quad l=2,\ldots,L\\
\mathbf{y} &= W^{(L)}\mathbf{v}^{(L)}.
\end{aligned}
\end{align}

Denote $C^{(l)}$ as the covariance of $\mathbf{v}^{(l)}$ and $(\tilde{W}^{(l)})_{l=0}^{L-1}$ as the weight matrices of the network. According to the chain rule, we have $\frac{\partial \mathcal{L}}{\partial  \tilde{W}_{ij}} = \frac{\partial \mathcal{L}}{\partial v^{(l+1)}_i }\frac{\partial v^{(l+1)}_i}{\partial  \tilde{W}_{ij}}$ for each layer $l$.
Thus, we can replace $\mathbf{x}$ in Eq.~\eqref{eq: label_y} with $\mathbf{v}^{(l)}$, replace $C_{\mathrm{in}}$ in Eq.~\eqref{eq: label_y2} with $C^{(l)}$, and replace $\tilde{C}_{\mathrm{in}}$ in Eq.~\eqref{eq: label_y} with $\tilde{C}^{(l)}$.
After applying these modifications, the rest of the analysis for each layer $l$ remains the same as the single-layer case.
Therefore, after trained an MNN $\tilde{M}$ with the same architecture, $\tilde{M}$ learns the weight matrices, means and variances for each layer of $M_{GT}$.

\subsection{MNNs approximate the covariance under mild conditions}\label{sec:cov_appro}
We analyze the covariance of the MNN $\tilde{M}$ after trained by SMUC. According to Eq.~\eqref{eq:heaviside_chi}, similar to the mean activation of Heaviside, we also have the scalar property
\begin{align}
    \chi(\mu,C)  = \chi(a\mu,a^2C) ,\quad \forall a> 0,\mu\in\mathbb{R},C>0
\end{align}
and hence
\begin{align}\label{eq:chi_right}
    \chi(\tilde{\mathbf{w}}_k^\mathrm{T}\mathbf{x},\tilde{C}_k) = \chi(\bar{\mu}_k,\bar{C}_k)
\end{align}
The covariance of $\tilde{M}$ is
\begin{align}
    \tilde{C}_{y,kj}(\mathbf{x}) 
=\chi(\bar{\mu}_k,\bar{C}_k)\chi(\bar{\mu}_j,\bar{C}_j) \frac{\mathbf{w}_k^T\tilde{C}_{\mathrm{in}}\mathbf{w}_j}{\sqrt{\bar{C}_k\bar{C}_j}} ,
    \quad k,j = 1,\ldots,n,\quad k\neq j.\label{eq:corr_error}
\end{align}
It is observed that unless $\mathbf{w}_k^T\tilde{C}_{\mathrm{in}}\mathbf{w}_j=\mathbf{w}_k^TC_{\mathrm{in}}\mathbf{w}_j$, we cannot ensure that the output covariance of $\tilde{M}$ is as the same as the ground-truth covariance $C_{y,kj}$ of $M_{GT}$.
When the dimension of the output is one ($n=1$), it is not an issue, since we only need to fit the variance of the output. When the dimension of the output is multi-dimensional ($n>1$), however, it seems that we need to know the ground-truth input covariance $C_{\mathrm{in}}$ for the learning the ground-truth output covariance.
Nevertheless, we show that under some mild conditions, when the number of the layers goes to infinity, the difference between the output covariance of $\tilde{M}$ and $M_{GT}$ goes to zero, i.e., we prove the Thm.~\ref{thm:cov}.

\begin{proof}
According to Eq.~\eqref{eq:chi_right}-\eqref{eq:corr_error}, we have
\begin{align}
    \left|C^{(l+1)}_{ij}-\tilde{C}^{(l+1)}_{ij}\right|
    &=  \left|\frac{\chi_i^{(l)}\chi_j^{(l)}}{\sqrt{\big(W_{i\bullet}^{(l)}C^{(l)}(W_{i\bullet}^{(l)})^{\mathrm{T}}+\sigma_l^2\big)\big(W_{j\bullet}^{(l)}C^{(l)}(W_{j\bullet}^{(l)})^{\mathrm{T}}+\sigma_l^2\big)}} \sum_{k,q}W^{(l)}_{ik}W^{(l)}_{jq}(C^{(l)}_{kq}-\tilde{C}^{(l)}_{kq})\right|
    \\
    &\leq \left|\chi_i^{(l)}\chi_j^{(l)}\right|\frac{n^{(l)}\left\|W_{i\bullet}^{(l)}\right\|_{2}\left\|W_{j\bullet}^{(l)}\right\|_{2}}{\sqrt{\big(\lambda_{\min}(C^{(l)})\left\|W_{i\bullet}^{(l)}\right\|_{2}^2+\sigma_l^2\big)\big(\lambda_{\min}(C^{(l)})\left\|W_{j\bullet}^{(l)}\right\|_{2}^2+\sigma_l^2\big)}}\max_{k,q}\left| C^{(l)}_{kq}-\tilde{C}^{(l)}_{kq}\right|\\
    &\leq \left|\chi_i^{(l)}\chi_j^{(l)}\right|\frac{n^{(l)}}{\sqrt{\lambda_{\min}(C^{(l)})^2+2\lambda_{\min}(C^{(l)})\sigma_l^2}}\max_{k,q}\left| C^{(l)}_{kq}-\tilde{C}^{(l)}_{kq}\right|,\quad i\neq j
\end{align}
where $\left\|\cdot\right\|_{2}$ is the $l_2$ norm of vectors. 
In the above, we have applied the Cauchy-Schwartz inequality
\begin{align}
\sum_{k,q}W^{(l)}_{ik}W^{(l)}_{jq}
    \leq n^{(l)}\left\|W_{i\bullet}^{(l)}\right\|_{2}\left\|W_{j\bullet}^{(l)}\right\|_{2},
\end{align}
and the arithmetic mean inequality
\begin{align}
  \frac{\left\|W_{i\bullet}^{(l)}\right\|_{2}^2}{\left\|W_{j\bullet}^{(l)}\right\|_{2}^2}+\frac{\left\|W_{j\bullet}^{(l)}\right\|_{2}^2}{\left\|W_{i\bullet}^{(l)}\right\|_{2}^2} \geq 2\sqrt{\frac{\left\|W_{i\bullet}^{(l)}\right\|_{2}^2}{\left\|W_{j\bullet}^{(l)}\right\|_{2}^2}\frac{\left\|W_{j\bullet}^{(l)}\right\|_{2}^2}{\left\|W_{i\bullet}^{(l)}\right\|_{2}^2}}=2
\end{align}
Denote the infinity norm of a vector or matrix as $\left\|\cdot\right\|_{\infty}$, then we have
\begin{align}\label{eq:inequ_error_cov}
    \left\|C^{(l+1)}-\tilde{C}^{(l+1)}\right\|_{\infty}\leq r^{(l)}\left\|C^{(l)}-\tilde{C}^{(l)}\right\|_{\infty},
\end{align}
where $ \left\|C^{(l)}-\tilde{C}^{(l)}\right\|_{\infty}$ is the the deviation of the trained network's covariance $\tilde{C}^{(l+1)}$ from the ground-truth one $C^{(l)}$ at the $l$-th layer, 
and $r^{(l)}$ is the {\em covariance deviation rate} at the $l$-layer defined in Eq.~\eqref{eq:dev_rate} in the main paper.
It suggests that $r^{(l)}$ describes how this deviation reduces through the $l$-layer.
If $-\sum_{l} \log r^{(l)}$ diverges to positive infinity, as assumed by the condition, we have:
\begin{align}\label{eq:goes_to_zero}
    \lim_{l \rightarrow +\infty}\left\|C^{(l)}_{ij}-\tilde{C}^{(l)}_{ij}\right\|_{\infty} \leq \lim_{l \rightarrow +\infty}\exp(\sum_{q=1}^l \log r^{(q)})\left\|C^{(1)}_{ij}-\tilde{C}^{(1)}_{ij}\right\|_{\infty} = 0.
\end{align}
\end{proof}
According to Eq.\eqref{eq:heaviside_chi} and Eq.\eqref{eq:inequ_error_cov}, to reduce the deviation in the output covariance, it is beneficial to have large $\frac{\mu^{(l)}}{\sqrt{\bar{C}^{(l)}}}$ and $\lambda_{\min}(C^{(l)})$ and have small $n^{(l)}$.
Note that $\frac{\mu^{(l)}}{\sqrt{\bar{C}^{(l)}}}$ is the signal-to-noise ratio (SNR) at the $l$-th layer, which reflects the quality of the inputs from the previous layer, it suggests that a high SNR helps to decrease the deviation of covariance.
Additionally, the amount $\lambda_{\min}(C^{(l)})$ is related to the fluctuation of the output covariance at the layer $l$. Accordingly, high output fluctuation may also benefits to the decreasing deviation of covariance when keeping $\frac{\mu^{(l)}}{\sqrt{\bar{C}^{(l)}}}$ as the same.
Due to the exponential term in the function $\chi$, the covariance rate $r^{(l)}$ exhibits an exponential decrease as the signal-to-noise ratio (SNR) increases. On the other hand, the covariance rate increases linearly with the dimension $n^{(l)}$. Consequently, even if the dimension $n^{(l)}$ is high, it is a moderately high SNR is sufficient to control the deviation rate.

\subsection{General activation functions}\label{sec:general_act}
The previous results require the scaling property of the MAs of (Eq.~\eqref{eq:scalar_property}), which may not be satisfied when the stochastic network applies general nonlinearities such as ReLU function. 
Nevertheless, we can generalize the above results for general activation function via {\it the simple function approximation theorem}~\cite{royden1988real}, which argues that
we can approximate any measurable nonlinearities $g$ by a series of shifted Heaviside functions i.e., $g_n(x)\rightarrow g(x)$ point-wisely as $n$ goes to infinity, where $  g_n(x)= \sum_{r=1}^{R_n}a_{r_n}h(x-b_{r_n})$
satisfying
\begin{align}\label{eq:g_abs_g_n}
    \left|g_n(x)\right|\leq \left|g(x)\right|,\quad\forall x\in\mathbb{R}, n\in\mathbb{N}
\end{align}
and $a_{r_n}$ and $b_{r_n}$ are constants.
This theorem implies that linear combination of the Heaviside functions with different biases $b_r$ can approximate any measurable activation function $g$. 
To demonstrate the validity of this approximation in training MNNs, specifically regarding the model's output and the calculation of the derivative of the mean activation with respect to the input mean during training, we need to verify
\begin{align}
    &\lim_{n\rightarrow +\infty}\int g_n(x)\mathcal{N}(x;\mu,\sigma^2)dx = \int g(x) \mathcal{N}(x;\mu,\sigma^2)dx,\quad \forall\mu,\sigma, \quad\text{(zero-order approximation),}\\
    &\lim_{n\rightarrow +\infty}\frac{d}{d\mu}\int g_n(x)\mathcal{N}(x;\mu,\sigma^2)dx = 
    \frac{d}{d\mu}\int g(x) \mathcal{N}(x;\mu,\sigma^2)dx,\quad \forall \mu,\sigma,\quad\text{(first-order approximation)},\label{eq:first_appro}
\end{align}
where we use $\mathcal{N}(x;\mu,\sigma^2)$
to represent the probabilistic density function of a Gaussian distribution with mean $\mu$ and variance $\sigma^2$.
We prove the above claims for the cases where the activation function $g$ satisfies that
\begin{align}\label{eq:cond_no_infy}
    \sup_{x\in\mathbb{R}}\frac{|g(x)|}{|x|^m+1} < +\infty 
\end{align}
for some $m>0$. This is a mild condition satisfied by most of common used activation functions.
The proof is as follows. For given $\mu$, $\sigma$, and $\epsilon>0$, we show that there exists $N(\epsilon)$, such that $\forall n>N(\epsilon)$, we have
\begin{align}
\left|\lim_{n\rightarrow +\infty}\int g_n(x)\mathcal{N}(x;\mu,\sigma^2)dx - \int g(x) \mathcal{N}(x;\mu,\sigma^2)dx\right|<\epsilon.
\end{align}
 Using Eq.~\eqref{eq:g_abs_g_n}, and the condition Eq.~\eqref{eq:cond_no_infy}, there exists a constant $A$ (independent of $n$) such that
 \begin{align}
   \frac{|g_n(x)-g(x)|}{|x|^m+1}
   \leq \frac{|g(x)|+|g_n(x)|}{|x|^m+1}\leq \frac{2|g(x)|}{|x|^m+1} \leq A.
 \end{align}
 Since $(|x|^m+1)\mathcal{N}(x;\mu,\sigma^2)$ is integrable, there exists $B>0$ such that
 \begin{align}
     A\int_{x \notin [-B,B]} (|x|^m+1)\mathcal{N}(x;\mu,\sigma^2)dx < \frac{\epsilon}{2}.
 \end{align}
 Since $[-B,B]$ is bounded, we can require $g_n$ to uniformly converges to $g$ on $[-B,B]$~\cite{royden1988real}. Hence, there exists a $N(\epsilon)$ such that
 \begin{align}
     \sup_{x\in[-B,B]}|g_n(x)-g(x)| <\frac{\epsilon}{2}.
 \end{align}
 Therefore, $\forall n>N(\epsilon)$, we have
\begin{align}
     &\left|\int g_n(x)\mathcal{N}(x;\mu,\sigma^2)dx - \int g(x) \mathcal{N}(x;\mu,\sigma^2)dx \right|\\
     &\leq \int |g_n(x)-g(x)|\mathcal{N}(x;\mu,\sigma^2)dx \\
     &< \frac{\epsilon}{2} \int_{x\in[-B,B]} \mathcal{N}(x;\mu,\sigma^2)dx + A\int_{x\notin[-B,B]} (|x|^m+1)\mathcal{N}(x;\mu,\sigma^2)dx <  \frac{\epsilon}{2}  + \frac{\epsilon}{2}=\epsilon 
\end{align}
The proof of the first-order approximation (Eq.~\eqref{eq:first_appro}) is as the same, except that we need to replace $\mathcal{N}(x;\mu,\sigma^2)$ by $\frac{d}{d\mu}\mathcal{N}(x;\mu,\sigma^2)$ in above procedure.
Hence, we can prove the approximation up to any finite order.
As a result, we can approximate an MNN that employs any measurable functions as activation functions with an MNN that utilizes the Heaviside function as the activation function. This approximation holds true not only for the model's output but also for the training process of the MNN.
In doing so, the previous theoretical results of SMUC can be directly generalized to the cases where MNN uses MAs derived from other nonlinearity such as ReLU MAs by approximating this MNN using corresponding Heaviside MNNs.

In contrast to MNNs, it is important to note that it is invalid to approximate a deterministic conventional ANNs using the Heaviside function as the activation function in terms of training. This is because the derivative of a linear combination of a finite number of Heaviside functions is almost zero everywhere, while the derivative of the activation function is generally non-zero at different points. 
In MNNs, the activation function's cumulants are smoothed by the Gaussian kernel. Consequently, if we can approximate the activation function up to the zeroth order, we can effectively approximate its cumulants up to any desired order

\end{document}